\documentclass{article}

\PassOptionsToPackage{numbers}{natbib}

\usepackage[final]{neurips_2018}

\usepackage[utf8]{inputenc} 
\usepackage[T1]{fontenc}    
\usepackage{hyperref}       
\usepackage{url}            
\usepackage{booktabs}       
\usepackage{amsfonts}       
\usepackage{nicefrac}       
\usepackage{microtype}      

\usepackage{smile}

\hypersetup{colorlinks,
linkcolor=red,
citecolor=blue,
urlcolor=magenta,
linktocpage,
plainpages=false}     

\usepackage{enumitem}
\usepackage{wrapfig,lipsum}

\usepackage{xcolor,colortbl}

\def \tO{\tilde{O}}

\newcommand{\ignore}[1]{}
\newcommand{\la}{\langle}
\newcommand{\ra}{\rangle}

\title{Global Convergence of Langevin Dynamics Based Algorithms for Nonconvex Optimization}

\author{Pan Xu\footnotemark[1] \\
  Department of Computer Science\\
  UCLA\\
  Los Angeles, CA 90095 \\
  \texttt{panxu@cs.ucla.edu} \\
  \And
  Jinghui Chen\thanks{Equal contribution.} \\
  Department of Computer Science \\
  University of Virginia\\
  Charlottesville, VA 22903 \\
  \texttt{jc4zg@virginia.edu} \\
  \AND
  Difan Zou \\
  Department of Computer Science\\
  UCLA\\
  Los Angeles, CA 90095 \\
  \texttt{knowzou@cs.ucla.edu} \\
  \And
  Quanquan Gu \\
  Department of Computer Science\\
  UCLA\\
  Los Angeles, CA 90095 \\
  \texttt{qgu@cs.ucla.edu}
}

\begin{document}

\maketitle

\begin{abstract}
We present a unified framework to analyze the global convergence of Langevin dynamics based algorithms for nonconvex finite-sum optimization with $n$ component functions.  At the core of our analysis is a direct analysis of the ergodicity of the numerical approximations to Langevin dynamics, which leads to faster convergence rates. Specifically, we show that gradient Langevin dynamics (GLD) and stochastic gradient Langevin dynamics (SGLD)  converge to the \textit{almost minimizer}\footnote{Following \cite{raginsky2017non}, an almost minimizer is defined to be a point which is within the ball of the global minimizer with radius $O(d\log(\beta+1)/\beta)$, where $d$ is the problem dimension and $\beta$ is the inverse temperature parameter.} within $\tilde O\big(nd/(\lambda\epsilon) \big)$ and $\tilde O\big(d^7/(\lambda^5\epsilon^5) \big)$ stochastic gradient evaluations respectively\footnote{$\tilde O(\cdot)$ notation hides polynomials of logarithmic terms and constants.}, where $d$ is the problem dimension, and $\lambda$ is the spectral gap of the Markov chain generated by GLD. Both results improve upon the best known gradient complexity\footnote{Gradient complexity is defined as the total number of stochastic gradient evaluations of an algorithm, which is the number of stochastic gradients calculated per iteration times the total number of iterations.} results \citep{raginsky2017non}. 
Furthermore, for the first time we prove the global convergence guarantee for variance reduced stochastic gradient Langevin dynamics (SVRG-LD) to the almost minimizer within $\tilde O\big(\sqrt{n}d^5/(\lambda^4\epsilon^{5/2})\big)$ stochastic gradient evaluations, which outperforms the gradient complexities of GLD and SGLD in a wide regime.  
Our theoretical analyses shed some light on using Langevin dynamics based algorithms for nonconvex optimization with provable guarantees.
\end{abstract}

\section{Introduction}
We consider the following nonconvex finite-sum optimization problem  
\begin{align}\label{def:opt_problem}
\min_{\xb} F_n(\xb):= \frac{1}{n}\sum_{i=1}^n f_i(\xb),
\end{align}
where $f_i(\xb)$'s are called component functions, and both $F_n(\xb)$ and $f_i(\cdot)$'s can be nonconvex. 
Various first-order optimization algorithms such as gradient descent \citep{nesterov2013introductory}, stochastic gradient descent \citep{ghadimi2013stochastic} and more recently variance-reduced stochastic gradient descent \citep{reddi2016stochastic, allen2016variance} have been proposed and analyzed for solving \eqref{def:opt_problem}. However, all these algorithms are only guaranteed to converge to a stationary point, which can be a local minimum, a local maximum, or even a saddle point. This raises an important question in nonconvex optimization and machine learning: is there an efficient algorithm that is guaranteed to converge to the global minimum of \eqref{def:opt_problem}? 


Recent studies by  \citet{dalalyan2014theoretical, dalalyan2017further} showed that sampling from a distribution which concentrates around the global minimum of $F_n(\xb)$ is a similar task as minimizing $F_n$ via certain optimization algorithms. This justifies the use of Langevin dynamics based algorithms for optimization. In detail, the first order Langevin dynamics is defined by the following stochastic differential equation (SDE)
\begin{align}\label{eq:langevin diffusion}
d\bX(t)= - \nabla F_n(\bX(t))dt + \sqrt{2 \beta^{-1} } d \bB(t),
\end{align}
where $\beta>0$ is the inverse temperature parameter that is treated as a constant throughout the analysis of this paper, and $\{\bB(t)\}_{t\geq0}$ is the standard Brownian motion in $\RR^d$.  
Under certain assumptions on the drift coefficient $\nabla F_n$, it was showed that the distribution of diffusion $\bX(t)$ in \eqref{eq:langevin diffusion} converges to its stationary distribution \citep{chiang1987diffusion}, a.k.a., the Gibbs measure $\pi(d\xb)\propto \exp(-\beta F_n(\xb))$, which concentrates on the global minimum of $F_n$ \citep{hwang1980laplace,gelfand1991recursive,roberts1996exponential}. Note that the above convergence result holds even when $F_n(\xb)$ is nonconvex. This motivates the use of Langevin dynamics based algorithms for nonconvex optimization \citep{raginsky2017non,zhang2017hitting,tzen2018local,simsekli2018asynchronous}. However, unlike first order optimization algorithms \citep{nesterov2013introductory,ghadimi2013stochastic,reddi2016stochastic,allen2016variance}, which have been extensively studied, the non-asymptotic theoretical guarantee of applying Langevin dynamics based algorithms for nonconvex optimization, is still under studied. 
In a seminal work, \citet{raginsky2017non} provided a non-asymptotic analysis of stochastic gradient Langevin dynamics (SGLD) \citep{welling2011bayesian} for nonconvex optimization, which is a stochastic gradient based discretization of \eqref{eq:langevin diffusion}. They proved that SGLD converges to an almost minimizer up to $d^2/(\sigma^{1/4}\lambda^*) \log(1/\epsilon)$ within $\tO(d/(\lambda^*\epsilon^4))$ iterations, where $\sigma^2$ is the variance of stochastic gradient and $\lambda^*$ is called the \textit{uniform spectral gap} of Langevin diffusion \eqref{eq:langevin diffusion}, 
and 
it is in the order of $e^{-\tO(d)}$. 
In a concurrent work, \citet{zhang2017hitting} analyzed the hitting time of SGLD and proved its convergence to an approximate local minimum. More recently, \citet{tzen2018local} studied the local optimality and generalization performance of Langevin algorithm for nonconvex functions through the lens of metastability and \citet{simsekli2018asynchronous} developed an asynchronous-parallel stochastic L-BFGS algorithm for non-convex optimization based on variants of SGLD. \citet{erdogdu2018global} further developed non-asymptotic analysis of global optimization based on a broader class of diffusions.  

In this paper, we establish the global convergence for a family of Langevin dynamics based algorithms, including Gradient Langevin Dynamics (GLD) \citep{dalalyan2014theoretical,durmus2015non,dalalyan2017further}, Stochastic Gradient Langevin Dynamics (SGLD) \citep{welling2011bayesian} and Stochastic Variance Reduced Gradient Langevin Dynamics (SVRG-LD) \citep{dubey2016variance} for solving the finite sum nonconvex optimization problem in \eqref{def:opt_problem}.  Our analysis is built upon the direct analysis of the discrete-time Markov chain rather than the continuous-time Langevin diffusion, and therefore avoid the discretization error. 

\subsection{Our Contributions}
The major contributions of our work are summarized as follows:
\begin{itemize}[leftmargin=*]
    \item We provide a unified analysis for a family of Langevin dynamics based algorithms by a new decomposition scheme of the optimization error, under which we directly analyze the ergodicity of numerical approximations for Langevin dynamics (see Figure \ref{fig:flowchart}). 
    
    \item Under our unified framework, we establish the global convergence of GLD for solving \eqref{def:opt_problem}. In detail, GLD requires $\tO\big(d/(\lambda\epsilon)\big)$ iterations to converge to the almost minimizer of \eqref{def:opt_problem} up to precision $\epsilon$, where $\lambda$ is the spectral gap of the discrete-time Markov chain generated by GLD and is in the order of $e^{-\tO(d)}$. This improves the $\tilde O\big(d/(\lambda^*\epsilon^4))$ iteration complexity of GLD implied by \cite{raginsky2017non}, where $\lambda^*=e^{-\tO(d)}$ is the spectral gap of Langevin diffusion \eqref{eq:langevin diffusion}.
    
    \item We establish a faster convergence of SGLD to the almost minimizer of \eqref{def:opt_problem}. In detail, it converges to the almost minimizer up to $\epsilon$ precision within $\tO\big({d^7}/{(\lambda^5\epsilon^5)} \big)$ stochastic gradient evaluations. 
    This also improves the $\tO\big(d^{17}/({\lambda^*}^8\epsilon^8)\big)$ gradient complexity proved in \cite{raginsky2017non}. 
     
    \item We also analyze the SVRG-LD algorithm and investigate its global convergence property. We show that SVRG-LD is guaranteed to converge to the almost minimizer of \eqref{def:opt_problem} within 
    $\tO\big( \sqrt{n}d^5/(\lambda^4\epsilon^{5/2}) \big) $ stochastic gradient evaluations. 
    It outperforms the gradient complexities of both GLD and SGLD when $1/\epsilon^3 \leq n \leq 1/\epsilon^5$.   
    To the best of our knowledge, this is the first global convergence guarantee of SVRG-LD for nonconvex optimization, while the original paper \citep{dubey2016variance} only analyzed the posterior sampling property of SVRG-LD. 
\end{itemize}

\subsection{Additional Related Work}
Stochastic gradient Langevin dynamics (SGLD)  \citep{welling2011bayesian} and its extensions \citep{ahn2012bayesian,ma2015complete,dubey2016variance} 
have been widely used in Bayesian learning. A large body of work has focused on analyzing the mean square error of Langevin dynamics based algorithms. In particular, \citet{vollmer2016exploration} analyzed the non-asymptotic bias and variance of the SGLD algorithm by using Poisson equations. \citet{chen2015convergence} showed the non-asymptotic bias and variance of MCMC algorithms with high order integrators.
\citet{dubey2016variance} proposed variance-reduced algorithms based on stochastic gradient Langevin dynamics, namely SVRG-LD and SAGA-LD, for Bayesian posterior inference, and proved that their method improves the mean square error upon SGLD. \citet{li2018stochastic} further improved the mean square error by applying the variance reduction
tricks on Hamiltonian Monte Carlo, which is also called  the underdamped Langevin dynamics. 

Another line of research \citep{dalalyan2014theoretical,durmus2016high,dalalyan2017further,dalalyan2017user,dwivedi2018log,zou2018stochastic} focused on characterizing the distance between distributions generated by Langevin dynamics based algorithms and (strongly) log-concave target distributions. In detail, \citet{dalalyan2014theoretical} proved that the distribution of the last step in GLD converges to the stationary distribution in $\tO(d/\epsilon^2)$ iterations in terms of total variation distance and Wasserstein distance respectively with a warm start and showed the similarities between posterior sampling and optimization. Later \citet{durmus2015non} improved the results by showing this result holds for any starting point and established similar bounds for the Wasserstein distance. \citet{dalalyan2017further} further improved the existing results in terms of the Wasserstein distance and provide further insights on the close relation between approximate sampling and gradient descent. \citet{cheng2018underdamped} improved existing 2-Wasserstein results by reducing the discretization error using underdamped Langevin dynamics. To improve the convergence rates in noisy gradient settings, \citet{chatterji2018theory,zou2018subsampled} presented convergence guarantees in 2-Wasserstein distance for SAGA-LD and SVRG-LD using variance reduction techniques. \citet{zou2018stochastic} proposed the variance reduced Hamilton Monte Carlo to accelerate the convergence of Langevin dynamics based sampling algorithms. As to sampling from distribution with compact support, \citet{bubeck2015sampling} analyzed sampling from log-concave distributions via projected Langevin Monte Carlo, and \citet{brosse2017sampling} proposed a proximal Langevin Monte Carlo algorithm. This line of research is orthogonal to our work since their analyses are regarding to the convergence of the distribution of the iterates to the stationary distribution of Langevin diffusion in total variation distance or 2-Wasserstein distance instead of expected function value gap. 

 
 On the other hand, many attempts have been made to escape from saddle points in nonconvex optimization, such as 
 cubic regularization \citep{nesterov2006cubic,pmlr-v80-zhou18d}, trust region Newton method \citep{curtis2014trust}, Hessian-vector product based methods \citep{agarwal2016finding,carmon2016gradient,carmon2016accelerated}, noisy gradient descent \citep{ge2015escaping, jin2017escape,jin2017accelerated} and normalized gradient \citep{levy2016power}. Yet all these algorithms are only guaranteed to converge to an approximate local minimum rather than a global minimum. The global convergence for nonconvex optimization remains understudied.

\subsection{Notation and Preliminaries}
In this section, we present notations used in this paper and some preliminaries for SDE. We use lower case bold symbol $\xb$ to denote deterministic vector, and use upper case italicized bold symbol $\bX$ to denote random vector. For a vector $\xb \in \RR^d$, we denote by $\|\xb\|_2$ its Euclidean norm. 
We use $a_n=O(b_n)$ to denote that  $a_n\leq C b_n$ for some constant $C>0$ independent of $n$. We also denote $a_n\lesssim b_n$ ($a_n\gtrsim b_n$) if $a_n$ is less than (larger than) $b_n$ up to a constant.
We also use $\tO(\cdot)$ notation to hide both polynomials of logarithmic terms and constants.

\noindent\textbf{Kolmogorov Operator and Infinitesimal Generator}\\
Suppose $\bX(t)$ is the solution to the diffusion process represented by the stochastic differential equation \eqref{eq:langevin diffusion}. For such a continuous time Markov process, let $P = \{P_t\}_{t>0}$ be the corresponding Markov semi-group \citep{bakry2013analysis}, and we define the Kolmogorov operator \citep{bakry2013analysis} $P_s$ as follows
\begin{align*}
    P_s g(\bX(t)) = \EE[g(\bX(s+t))|\bX(t)],
\end{align*}
where $g$ is a smooth test function.
We have $P_{s+t} = P_s \circ P_t$ by Markov property.
Further we define the infinitesimal generator \citep{bakry2013analysis} of the semi-group $\cL$ to describe the the movement of the process in an infinitesimal time interval:
\begin{align*}
    \cL g(\bX(t)) :&= \lim_{h \rightarrow 0^+} \frac{\EE[g(\bX(t+h))|\bX(t)] - g(\bX(t))}{h} \\
    &= \big( -\nabla F_n(\bX(t))\cdot \nabla +  \beta^{-1} \nabla^2\big)g(\bX(t)),
\end{align*}
where $\beta$ is the inverse temperature parameter.

\noindent\textbf{Poisson Equation and the Time Average}\\
Poisson equations are widely used in the study of homogenization and ergodic theory to prove the desired limit of a time-average. Let $\cL$ be the infinitesimal generator and let $\psi$ be defined as follows
\begin{align}\label{eq:poisson_equation}
    \cL \psi = g - \bar g,
\end{align}
where $g$ is a smooth test function and $\bar g$ is the expectation of $g$ over the Gibbs measure, i.e., $\bar g:=\int g(\xb)\pi(d\xb)$. Smooth function $\psi$ is called the solution of Poisson equation \eqref{eq:poisson_equation}. Importantly, it has been shown \cite{erdogdu2018global} that the first and second order derivatives of the solution $\psi$ of Poisson equation for Langevin diffusion can be bounded by polynomial growth functions.

\section{Review of Langevin Dynamics Based Algorithms}\label{sec:alg}

In this section, we briefly review three Langevin dynamics based algorithms proposed recently.



In practice, numerical methods (a.k.a., numerical integrators) are used to approximate the Langevin diffusion in \eqref{eq:langevin diffusion}. For example, by Euler-Maruyama scheme \citep{kloeden1992higher},  \eqref{eq:langevin diffusion} can be discretized as follows:
\begin{align}\label{eq:gld}
\bX_{k+1} = \bX_k - \eta \nabla F_n(\bX_k) + \sqrt{2\eta \beta^{-1} } \cdot\bepsilon_k,
\end{align}
where $\bepsilon_k \in \RR^d$ is standard Gaussian noise and $\eta>0$ is the step size. The update in \eqref{eq:gld} resembles gradient descent update except for an additional injected Gaussian noise. The magnitude of the Gaussian noise is controlled by the inverse temperature parameter $\beta$. In our paper, we refer this update as Gradient Langevin Dynamics (GLD) \citep{dalalyan2014theoretical,durmus2015non,dalalyan2017further}. The details of GLD are shown in Algorithm \ref{alg:gld}.

In the case that $n$ is large, the above Euler approximation can be infeasible due to the high computational cost of the full gradient $\nabla F_n(\bX_k)$ at each iteration. A natural idea is to use stochastic gradient to approximate the full gradient, which gives rise to Stochastic Gradient Langevin Dynamics (SGLD) \citep{welling2011bayesian} and its variants \citep{ahn2012bayesian,ma2015complete,chen2015convergence}. 
However, the high variance brought by the stochastic gradient can make the convergence of SGLD slow. To reduce the variance of the stochastic gradient and accelerate the convergence of SGLD, we use a mini-batch of stochastic gradients in the following update form:
\begin{align}\label{eq:sgld}
\bY_{k+1} = \bY_k - \frac{\eta}{B} \sum_{i\in I_k} \nabla f_{i}(\bY_k) + \sqrt{2\eta \beta^{-1} } \cdot\bepsilon_k,
\end{align}
where $1/B\sum_{i\in I_k}\nabla f_{i}(\bY_k)$ is the stochastic gradient, which is an unbiased estimator for $\nabla F_n(\bY_k)$ and $I_k$ is a subset of $\{1,\ldots,n\}$ with $|I_k|=B$. Algorithm \ref{alg:sgld} displays the details of SGLD.

Motivated by recent advances in stochastic optimization, in particular, the variance reduction based techniques \citep{johnson2013accelerating, reddi2016stochastic,allen2016variance},
\citet{dubey2016variance} proposed the Stochastic Variance Reduced Gradient Langevin Dynamics (SVRG-LD) for posterior sampling.
The key idea is to use semi-stochastic gradient to reduce the variance of the stochastic gradient. SVRG-LD takes the following update form:
\begin{align}\label{eq:SVRG-LD}
\bZ_{k+1} = \bZ_k - \eta \tilde\nabla_k + \sqrt{2\eta \beta^{-1} } \cdot\bepsilon_k, 
\end{align}
where $\tilde{\nabla}_{k}= 1/B \sum_{i_k \in I_k} \big(\nabla f_{i_k}(\bZ_k)-\nabla f_{i_k}(\tilde{\bZ}^{(s)})+ \nabla F_n(\tilde\bZ^{(s)})\big)$ is the semi-stochastic gradient, $\tilde\bZ^{(s)}$ is a snapshot of $\bZ_k$ at every $L$ iteration such that $k=sL+\ell$ for some $\ell=0,1,\ldots,L-1$, and $I_k$ is a subset of $\{1,\ldots,n\}$ with $|I_k|=B$. 
SVRG-LD is summarized in Algorithm \ref{alg:vr_sgld}.

Note that although all the three algorithms are originally proposed for posterior sampling or more generally, Bayesian learning, they can be applied for nonconvex optimization, as demonstrated in many previous studies \citep{ahn2012bayesian, raginsky2017non,zhang2017hitting}.

\begin{algorithm}[htp!]
	\caption{Gradient Langevin Dynamics (GLD)}
	\label{alg:gld}
	\begin{algorithmic}
		\STATE \textbf{input:} step size $\eta>0$; inverse temperature parameter $\beta>0$;
		$\bX_0 = \zero$
 		\FOR {$k = 0,1,\ldots, K-1$}
		     \STATE randomly draw $\bepsilon_k\sim N(\zero,\Ib_{d\times d})$
		     \STATE $\bX_{k+1}=\bX_k-\eta \nabla F_n(\bX_{k})+\sqrt{2\eta/\beta}\bepsilon_k$
		\ENDFOR     
	\end{algorithmic}
\end{algorithm}


\begin{algorithm}[ht]
	\caption{Stochastic Gradient Langevin Dynamics (SGLD)}
	\label{alg:sgld}
	\begin{algorithmic}
		\STATE \textbf{input:} step size $\eta>0$; batch size $B$; inverse temperature parameter $\beta>0$;
		$\bY_0 = \zero$
 		\FOR {$k = 0,1,\ldots, K-1$}
		     \STATE randomly pick a subset $I_k$ from $\{1,\ldots,n\}$ of size $|I_k|=B$; randomly draw $\bepsilon_k\sim N(\zero,\Ib_{d\times d})$
		     \STATE $\bY_{k+1}=\bY_k-\eta/B\sum_{i\in I_k} \nabla f_{i}(\bY_{k})+\sqrt{2\eta/\beta}\bepsilon_k$
		\ENDFOR     
	\end{algorithmic}
\end{algorithm}
 
\begin{algorithm}[htp!]
	\caption{Stochastic Variance Reduced Gradient Langevin Dynamics (SVRG-LD)}
	\label{alg:vr_sgld}
	\begin{algorithmic}
		\STATE \textbf{input:} step size $\eta>0$; batch size $B$; epoch length $L$; inverse temperature parameter $\beta>0$
		\STATE \textbf{initialization:} $\bZ_0 = \zero$, $\tilde{\bZ}^{(0)}=\bZ_0$
		\FOR {$s = 0,1,\ldots, (K/L)-1$}
		    \STATE $\tilde\bW = \nabla F_n(\tilde{\bZ}^{(s)})$
		    \FOR{$\ell = 0, \dots, L-1$ }
		    \STATE $k = s L+ \ell$
		    \STATE randomly pick a subset $I_k$ from $\{1,\ldots,n\}$ of size $|I_k|=B$; draw $\bepsilon_{k}\sim N(\zero,\Ib_{d\times d})$
		    \STATE $\tilde{\nabla}_{ k}= 1/B\sum_{i_k \in I_k} \big(\nabla f_{i_k}(\bZ_k)-\nabla f_{i_k}(\tilde{\bZ}^{(s)})+\tilde\bW\big)$
			\STATE $\bZ_{k+1}=\bZ_k-\eta\tilde{\nabla}_{k}+\sqrt{2\eta/\beta}\bepsilon_{k}$
			\ENDFOR
			\STATE
			$\tilde{\bZ}^{(s)}=\bZ_{(s+1)L}$
		\ENDFOR
	\end{algorithmic}
\end{algorithm}

 
\section{Main Theory}\label{sec:main_result}

Before we present our main results, we first lay out the following assumptions on the loss function. 



\begin{assumption}[Smoothness]\label{assump:smooth}
The function $f_i(\xb)$ is $M$-smooth for $M>0$, $i = 1,\dots, n$, i.e.,
\begin{align*}
    \|\nabla f_i(\xb)-\nabla f_i(\yb)\|_2\leq M\|\xb-\yb\|_2, \quad\text{ for any } \xb,\yb\in\RR^d.
\end{align*}
\end{assumption}
Assumption \ref{assump:smooth} immediately implies that $F_n(\xb) = 1/n\sum_{i=1}^n f_i(\xb)$ is also $M$-smooth. 
\begin{assumption}[Dissipative]\label{assump:dissipative}
There exist constants $m,b>0$, such that we have
$$\la \nabla F_n(\xb),\xb\ra\geq m\|\xb\|_2^2-b, \quad\text{ for all }\xb\in\RR^d.$$
\end{assumption}
Assumption \ref{assump:dissipative} is a typical assumption for the convergence analysis of an SDE and diffusion approximation \citep{mattingly2002ergodicity,raginsky2017non,zhang2017hitting}, which can be satisfied by enforcing a weight decay regularization \cite{raginsky2017non}. It says that starting from a position that is sufficiently far away from the origin, the Markov process defined by \eqref{eq:langevin diffusion} moves towards the origin on average. It can also be noted that all critical points are within the ball of radius $O(\sqrt{b/m})$ centered at the origin under this assumption. 


Let $\xb^*=\argmin_{\xb\in\RR^d} F_n(\xb)$ be the global minimizer of $F_n$. Our ultimate goal is to prove the convergence of the optimization error in expectation, i.e., $\EE[F_n(\bX_k)]-F_n(\xb^*)$. In the sequel, we decompose the optimization error into two parts: (1) $\EE[F_n(\bX_k)]-\EE[F_n(\bX^{\pi})]$, which characterizes the gap between the expected function value at the $k$-th iterate $\bX_k$ and the expected function value at $\bX^\pi$, where $\bX^{\pi}$ follows the stationary distribution $\pi(d\xb)$ of Markov process $\{\bX(t)\}_{t\geq0}$, and (2) $\EE[F_n(\bX^{\pi})]-F_n(\xb^*)$. Note that the error in part (1) is algorithm dependent, while that in part (2) only depends on the diffusion itself and hence is identical for all Langevin dynamics based algorithms.

Now we are ready to present our main results regarding to the optimization error of each algorithm reviewed in Section \ref{sec:alg}. We first show the optimization error bound of GLD (Algorithm \ref{alg:gld}).

\begin{theorem}[GLD]\label{thm:converge_gld}
Under Assumptions \ref{assump:smooth} and \ref{assump:dissipative}, consider $\bX_K$ generated by Algorithm \ref{alg:gld} with initial point $\bX_0=\zero$. The optimization error is bounded by
\begin{align}\label{eq:err_bound_gld}
    \EE[F_n(\bX_K)]-F_n(\xb^*) \leq \Theta e^{-\lambda K\eta}+ \frac{C_{\psi}\eta}{\beta}+\underbrace{\frac{d}{2\beta}\log\bigg(\frac{eM(b\beta/d+1)}{m}\bigg)}_{\cR_M},
\end{align}
where problem-dependent parameters $\Theta$ and $\lambda$ are defined as
\begin{align*}
    \Theta =\frac{C_0M(b\beta+m\beta+d)(m+e^{m\eta}M(b\beta+m\beta+d))}{m^2\rho_{\beta}^{d/2}}, \quad \lambda=\frac{2m\rho_{\beta}^d}{\log(2M(b\beta+m\beta+d)/m)},
\end{align*}
$C_0, C_{\psi}>0$ are constants, and $\rho_{\beta}\in(0,1)$ is a contraction parameter depending on the inverse temperature of the Langevin dynamics.
\end{theorem}

In the optimization error of GLD \eqref{eq:err_bound_gld}, we denote the upper bound of the error term $\EE[F_n(\bX^{\pi})]-F_n(\xb^*)$ by $\cR_M$, which characterizes the distance between the expected function value at $\bX^{\pi}$
and the global minimum of $F_n$. The stationary distribution of Langevin diffusion $\pi\propto e^{-\beta F_n(\xb)}$ is a Gibbs distribution, which concentrates around the minimizer $\xb^*$ of $F_n$. Thus a random vector $\bX^{\pi}$ following the law of $\pi$ is called an \textit{almost minimizer} of $F_n$ within a neighborhood of $\xb^*$ with radius $\cR_M$ \citep{raginsky2017non}. 

It is worth noting that the first term in \eqref{eq:err_bound_gld} vanishes at a exponential rate due to the ergodicity of Markov chain $\{\bX_{k}\}_{k=0,1\ldots}$. Moreover, the exponential convergence rate is controlled by $\lambda$, the spectral gap of the discrete-time Markov chain generated by GLD, which is in the order of $e^{-\tO(d)}$.

By setting $\EE[F_n(\bX_K)]-\EE[F_n(\bX^{\pi})]$ to be less than a precision $\epsilon$, and solving for $K$, we have the following corollary on the iteration complexity for GLD to converge to the almost minimizer $\bX^{\pi}$.
\begin{corollary}[GLD]\label{coro:gld}
Under the same conditions as in Theorem \ref{thm:converge_gld}, provided that $\eta  \lesssim \epsilon$,
GLD achieves $\EE[F_n(\bX_K)]-\EE[F_n(\bX^{\pi})]\leq\epsilon$
after
\begin{align*}
  K=O\bigg(\frac{d}{\epsilon\lambda}\cdot\log\bigg( \frac{1}{\epsilon }\bigg)\bigg)  
\end{align*}
iterations.
\end{corollary}

\begin{remark}
In a seminal work by \citet{raginsky2017non}, they provided a non-asymptotic analysis of SGLD for nonconvex optimization. By setting the variance of stochastic gradient to $0$, their result immediately suggests an $O(d/(\epsilon^4\lambda^*)\log^5((1/\epsilon)))$ iteration complexity for GLD to converge to the almost minimizer up to precision $\epsilon$.
Here the quantity $\lambda^*$ is the so-called \textbf{uniform spectral gap} for continuous-time Markov process $\{\bX_t\}_{t\geq 0}$ generated by Langevin dynamics. They further proved that $\lambda^*=e^{-\tO(d)}$, which is in the same order of our spectral gap $\lambda$ for the discrete-time Markov chain $\{\bX_{k}\}_{k=0,1\ldots}$ generated by GLD. Both of them match the lower bound for metastable exit times of SDE for nonconvex functions that have multiple local minima and saddle points \citep{bovier2004metastability}. Although for some specific function $F_n$, the spectral gap may be reduced to polynomial in $d$ \citep{ge2017beyond}, in general, the spectral gap for continuous-time Markov processes is in the same order as the spectral gap for discrete-time Markov chains. 
Thus, the iteration complexity of GLD suggested by Corollary \ref{coro:gld} is better than that suggested by \cite{raginsky2017non} by a factor of $O(1/\epsilon^3)$. 
\end{remark}


We now present the following theorem, which states the optimization error of SGLD (Algorithm \ref{alg:sgld}).
\begin{theorem}[SGLD]\label{thm:converge_sgld}
Under Assumptions \ref{assump:smooth} and \ref{assump:dissipative}, consider $\bY_K$ generated by Algorithm \ref{alg:sgld} with initial point $\bY_0=\zero$, the optimization error is bounded by
\begin{align}\label{eq:err_bound_sgld}
    &\EE[F_n(\bY_K)]-F_n(\xb^*)
    \leq  C_1\Gamma K\eta  \bigg[ \frac{\beta (n-B)(M\sqrt{\Gamma}+G)^2}{B(n-1)}  \bigg]^{1/2}+\Theta e^{-\lambda K\eta}+ \frac{C_{\psi}\eta}{\beta} +\cR_M, 
\end{align}
where $C_1$ is an absolute constant, $C_{\psi}, \lambda, \Theta $ and $\cR_M$ are the same as in Theorem \ref{thm:converge_gld}, $B$ is the mini-batch size, $G = \max_{i=1,\ldots,n}\{\|\nabla 
f_i(\xb^*)\|_2\} + bM/m$ and $\Gamma=2(1+1/m)(b+2G^2+d/\beta)$.
\end{theorem}
Similar to Corollary \ref{coro:gld}, by setting $\EE[F_n(\bY_k)]-\EE[F_n(\bX^{\pi})]\leq\epsilon$, we obtain the following corollary.  
\begin{corollary}[SGLD]\label{coro:sgld}
Under the same conditions as in Theorem \ref{thm:converge_sgld}, if $\eta \lesssim \epsilon$, 
SGLD achieves 
\begin{align}\label{eq:sgld_non_convergence}
\EE[F_n(\bY_K)]-\EE[F_n(\bX^{\pi})] = O\bigg(\frac{d^{3/2}}{B^{1/4}\lambda^{}}\cdot\log \bigg(\frac{1}{\epsilon}\bigg)+\epsilon\bigg),
\end{align}
after 
\begin{align*}
K=O\bigg(\frac{d}{\epsilon\lambda}\cdot \log\bigg(\frac{1}{\epsilon }\bigg)\bigg)
\end{align*}
iterations, where $B$ is the mini-batch size of Algorithm \ref{alg:sgld}.
\end{corollary}



\begin{remark}
Corollary \ref{coro:sgld} suggests that if the mini-batch size $B$ is chosen to be large enough to offset the divergent term $\log(1/\epsilon)$, SGLD is able to converge to the almost minimizer in terms of expected function value gap. 
This is also suggested by the result in \cite{raginsky2017non}. More specifically, 
the result in \cite{raginsky2017non} implies that SGLD achieves
\begin{align*}
    \EE[F_n(\bY_K)]-\EE[F_n(\bX^{\pi})]=O\bigg(\frac{d^2}{\lambda^{*}}  \bigg(\sigma^{-1/4}\log \bigg(\frac{1}{\epsilon}\bigg)+\epsilon\bigg)\bigg)
\end{align*}
with $K=O(d/(\lambda^*\epsilon^4)\cdot\log^5(1/\epsilon))$, where $\sigma^2$ is the upper bound of stochastic variance in SGLD, which can be reduced with larger batch size $B$. Recall that the spectral gap $\lambda^*$ in their work scales as $O(e^{-\tO(d)})$, which is in the same order as $\lambda$ in Corollary \ref{coro:sgld}. In comparison, our result in Corollary \ref{coro:sgld} indicates that SGLD can actually achieve the same order of error for $\EE[F_n(\bY_K)]-\EE[F_n(\bX^{\pi})]$ with substantially fewer number of  iterations, i.e., $O(d/(\lambda\epsilon)\log(1/\epsilon))$ .
\end{remark}

\begin{remark}
To ensure SGLD converges in Corollary \ref{coro:sgld}, one may set a sufficiently large batch size $B$ to offset the divergent term. For example, if we choose 
\begin{align*}
B\gtrsim \frac{d^6}{\lambda^{4}\epsilon^{4}}\log^{4}\bigg(\frac{1}{\epsilon}\bigg),    
\end{align*}
SGLD achieves $\EE[F_n(\bY_K)]-\EE[F_n(\bX^{\pi})]\leq\epsilon$ within $K=O(d/(\lambda\epsilon)\log(1/\epsilon))$ stochastic gradient evaluations. 
\end{remark}

\noindent In what follows, we proceed to present our result on the optimization error bound of SVRG-LD.
\begin{theorem}[SVRG-LD]\label{thm:converge_vrsgld}
Under Assumptions \ref{assump:smooth} and \ref{assump:dissipative}, consider  $\bZ_K$ generated by Algorithm \ref{alg:vr_sgld} with initial point $\bZ_0=\zero$. The optimization error is bounded by
\begin{align}\label{eq:err_bound_vr_sgld}
    &\EE[F_n(\bZ_K)]-F_n(\xb^*)\nonumber\\
    &\leq  
    C_1\Gamma K^{3/4}\eta \bigg[ \frac{L\beta M^2(n-B)}{B(n-1)}\bigg( 9\eta L(M^2 \Gamma+G^2) + \frac{ d}{\beta}\bigg)\bigg]^{1/4}+\Theta e^{-\lambda K\eta}+ \frac{C_{\psi}\eta}{\beta}+\cR_M,  
\end{align}
where constants $C_1,C_{\psi},\lambda,\Theta ,\Gamma,G$ and $\cR_M$ are the same as in Theorem \ref{thm:converge_sgld}, $B$ is the mini-batch size and $L$ is the length of inner loop of Algorithm \ref{alg:vr_sgld}.
\end{theorem}

Similar to Corollaries \ref{coro:gld} and \ref{coro:sgld}, we have the following iteration complexity for SVRG-LD.
\begin{corollary}[SVRG-LD]\label{coro:vr_sgld}
Under the same conditions as in Theorem \ref{thm:converge_vrsgld}, if $\eta \lesssim \epsilon$, SVRG-LD achieves $\EE[F_n(\bZ_K)]-\EE[F_n(\bX^{\pi})]\leq\epsilon$ after
$$K=O\bigg( \frac{Ld^5}{B\lambda^4\epsilon^4}\cdot\log^4\bigg(\frac{1}{\epsilon}\bigg)+ \frac{1}{\epsilon}\bigg)$$ 
iterations. 
In addition,
if we choose $B =  \sqrt{n}\epsilon^{-3/2}$, $L =  \sqrt{n}\epsilon^{3/2}$,
the number of stochastic gradient evaluations needed for SVRG-LD to achieve $\epsilon$ precision is 
\begin{align*}
    \tO\bigg( \frac{\sqrt{n}}{\epsilon^{5/2}}   \bigg)\cdot e^{\tO(d)}.
\end{align*}




\end{corollary}

\begin{remark}
In Theorem \ref{thm:converge_vrsgld} and Corollary \ref{coro:vr_sgld}, we establish the global convergence guarantee for SVRG-LD to an almost minimizer of a nonconvex function $F_n$. To the best of our knowledge, this is the first iteration/gradient complexity guarantee for SVRG-LD in nonconvex finite-sum optimization. \citet{dubey2016variance} first proposed the SVRG-LD algorithm for posterior sampling, but only proved that the mean square error between averaged sample pass and the stationary distribution
converges to $\epsilon$ within $\tO(1/\epsilon^{3/2})$ iterations, which has no implication for nonconvex optimization.  
\end{remark}
 
\begin{table}[h]
	\centering
	\caption{Gradient complexities of GLD, SGLD and SVRG-LD 
	to converge to the almost minimizer.
	\label{table:gradient_complexity}}
\begin{tabular}{cccc}
    \toprule
    &GLD&SGLD\footnotemark&SVRG-LD\\
	\midrule
	\cite{raginsky2017non} & $\tO\big(\frac{n}{\epsilon^4}\big)\cdot e^{\tO(d)}$& $\tO\big(\frac{1}{\epsilon^8}\big)\cdot e^{\tO(d)}$  & N/A\\
	This paper & $\tO\big(\frac{n}{\epsilon}\big)\cdot e^{\tO(d)}$&  $\tO\big(\frac{1}{\epsilon^5}\big)\cdot e^{\tO(d)}$ &$ \tO\Big( \frac{\sqrt{n}}{\epsilon^{5/2}} \Big)\cdot e^{\tO(d)} $\\
	\bottomrule
\end{tabular}
\end{table}
In large scale machine learning problems, the evaluation of full gradient can be  quite expensive, in which case the iteration complexity is no longer appropriate to reflect the efficiency of different algorithms. To perform a comprehensive comparison among the three algorithms, we present their gradient complexities for converging to the almost minimizer $\bX^{\pi}$ with $\epsilon$ precision in Table \ref{table:gradient_complexity}. 
Recall that gradient complexity is defined as the total number of stochastic gradient evaluations needed to achieve $\epsilon$ precision.
It can be seen from Table \ref{table:gradient_complexity} that the gradient complexity for GLD has worse dependence on the number of component functions $n$ and SVRG-LD has worse dependence on the optimization precision $\epsilon$. More specifically, when the number of component functions satisfies $n \leq 1/\epsilon^5$, SVRG-LD achieves better gradient complexity than SGLD. Additionally, if $n \geq 1/\epsilon^3$, SVRG-LD is better than both GLD and SGLD, therefore is more favorable.

\footnotetext{For SGLD in \cite{raginsky2017non}, the result in the table is obtained by choosing the exact batch size suggested by the authors that could make the stochastic variance small enough to cancel out the divergent term in their paper.} 

\section{Proof Sketch of the Main Results}\label{sec:proof_sketch}
In this section, we highlight our high level idea in the analysis of GLD, SGLD and SVRG-LD.

\subsection{Roadmap of the Proof}

Recall the problem in \eqref{def:opt_problem} and denote the global minimizer as $\xb^*=\argmin_{\xb}F_n(\xb)$.  $\{\bX(t)\}_{t\geq 0}$ and $\{\bX_k\}_{k=0,\ldots,K}$ are the continuous-time and discrete-time Markov processes generated by Langevin diffusion \eqref{eq:langevin diffusion} and GLD respectively. We propose to decompose the optimization error as follows:
\begin{align}\label{eq:decomp_excess_risk}
    &\EE [F_n(\bX_k)]-F_n(\xb^*)\notag\\
    &=\underbrace{\EE[F_n(\bX_k)]-\EE[F_n(\bX^{\mu})]}_{I_1}+\underbrace{\EE[F_n(\bX^{\mu})]-\EE[F_n(\bX^{\pi})]}_{I_2}+\underbrace{\EE[F_n(\bX^{\pi})]-F_n(\xb^*)}_{I_3},
\end{align}
where $\bX^{\mu}$ follows the stationary distribution $\mu(d\xb)$ of Markov process $\{\bX_k\}_{k=0,1,\ldots, K}$, and $\bX^{\pi}$ follows the stationary distribution $\pi(d\xb)$ of Markov process $\{\bX(t)\}_{t\geq 0}$, a.k.a.,  the Gibbs distribution. 
Following existing literature \citep{mattingly2002ergodicity,mattingly2010convergence,chen2015convergence}, here we assume the existence of stationary distributions, i.e., the ergodicity, of Langevin diffusion \eqref{eq:langevin diffusion} and its numerical approximation \eqref{eq:sgld}. Note that the ergodicity property of an SDE is not trivially guaranteed in general and establishing the existence of the stationary distribution is beyond the scope of our paper. Yet we will discuss the circumstances when geometric ergodicity holds in the Appendix. 

\begin{wrapfigure}{r}{4.60cm}
	\includegraphics[width=0.3\textwidth]{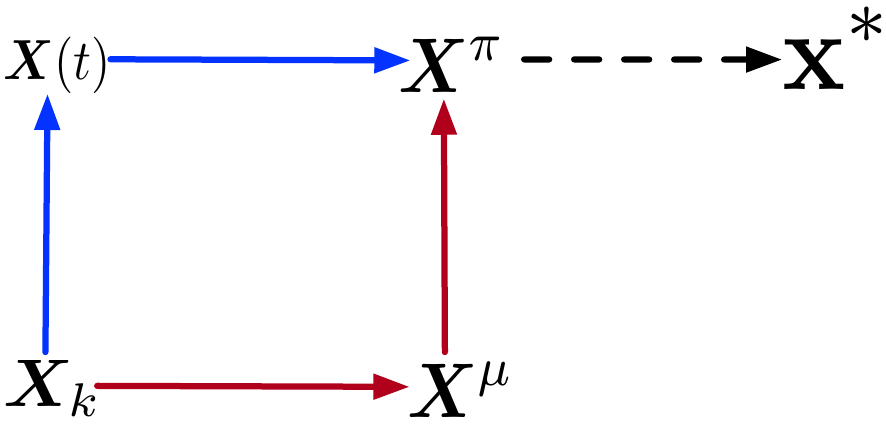}
	\caption{Illustration of the analysis framework in our paper. 
	}
	\label{fig:flowchart}
\end{wrapfigure}
We illustrate the decomposition  \eqref{eq:decomp_excess_risk} in Figure \ref{fig:flowchart}. Unlike existing optimization analysis of SGLD such as \cite{raginsky2017non}, which measure the approximation error between $\bX_k$ and $\bX(t)$ (blue arrows in the chart), we directly analyze the geometric convergence of discretized Markov chain $\bX_k$ to its stationary distribution (red arrows in the chart). Since the distance between $\bX_k$ and $\bX(t)$ is a slow-convergence term in \cite{raginsky2017non}, and the distance between $\bX(t)$ and $\bX^\pi$ depends on the uniform spectral gap, our new roadmap of proof will bypass both of these two terms, hence leads to a faster convergence rate. 

\noindent\textbf{Bounding $I_1$: Geometric Ergodicity of GLD}

To bound the first term in \eqref{eq:decomp_excess_risk}, we need to analyze the convergence of the Markov chain generated by Algorithm \ref{alg:gld} to its stationary distribution, namely, the ergodic property of the numerical approximation of Langevin dynamics. In probability theory, ergodicity  describes the long time behavior of Markov processes. For a finite-state Markov Chain, this is also closely related to the mixing time and has been thoroughly studied  in the literature of Markov processes \citep{hairer2008spectral,levin2009markov,bakry2013analysis}. Note that \citet{durmus2016high} studied the convergence of the Euler-Maruyama discretization (also referred to as the unadjusted Langevin algorithm) towards its stationary distribution in total variation. Nevertheless, they only focus on strongly convex functions which are less challenging than our nonconvex setting.

The following lemma ensures the geometric ergodicity of gradient Langevin dynamics.

\begin{lemma}\label{lemma:ergodic_gld}
Under Assumptions \ref{assump:smooth} and \ref{assump:dissipative}, the gradient Langevin dynamics (GLD) in Algorithm \ref{alg:gld} has a unique invariant measure $\mu$ on $\RR^d$.
It holds that
\begin{align*}
    |\EE[F_n(\bX_k)]-\EE[F_n(\bX^\mu)]| \leq C\kappa\rho_{\beta}^{-d/2}(1+\kappa e^{m\eta})\exp\bigg(-\frac{2mk\eta \rho_{\beta}^d}{\log(\kappa)}\bigg),
\end{align*}
where $C>0$ is an absolute constant, $\rho_{\beta}\in(0,1)$ is a contraction parameter depending on the inverse temperature parameter of Langevin dynamics \eqref{eq:langevin diffusion},  and $\kappa=2M(b\beta+m\beta+d)/b$.
\end{lemma} 
Lemma \ref{lemma:ergodic_gld} establishes the exponential decay of function gap between $F_n(\bX_k)$ and $F_n(\bX^{\pi})$ using coupling techniques. Note that the exponential dependence on dimension $d$ is consistent with the result from \cite{raginsky2017non} using entropy methods. The contraction parameter $\rho_{\beta}$ is a lower bound of the minorization condition for the Markov chain $\bX_k$, which is established in \cite{mattingly2002ergodicity}. Nonetheless, we would like to point out that the exact computation of $\rho_{\beta}$ requires additionally nontrivial efforts, which is beyond the scope of this work.

\noindent\textbf{Bounding $I_2$: Convergence to Stationary Distribution of Langevin Diffusion}

Now we are going to bound the distance between two invariant measures $\mu$ and $\pi$ in terms of their expectations over the objective function $F_n$. Our proof is inspired by that of  \citet{vollmer2016exploration,chen2015convergence}.  
The key insight here is that after establishing the geometric ergodicity of GLD, by the stationarity of $\mu$, we have
\begin{align*}
    \int F_n(\xb)\mu(d\xb) = \int \EE[F_n(\bX_k)|\bX_0 = \xb]\cdot\mu(d\xb).
\end{align*}
This property says that after reaching the stationary distribution, any further transition (GLD update) will not change the distribution. Thus we can bound the difference between two invariant measures.

\begin{lemma}\label{theorem:density_dist}
Under Assumptions \ref{assump:smooth} and \ref{assump:dissipative}, the invariant measures $\mu$ and $\pi$ satisfy
\begin{align*} 
    \big|\EE[F_n(\bX^{\mu})]-\EE[F_n(\bX^{\pi})]\big|
    &\leq \frac{C_{\psi}\eta}{\beta},
\end{align*}
where $C_{\psi}>0$ is a constant that dominates $\EE[\|\nabla^{p}\psi(\bX_k)\|]$ ($p=0,1,2$) and $\psi$ is the solution of Poisson equation \eqref{eq:poisson_equation}.
\end{lemma}
Lemma \ref{theorem:density_dist} suggests that the bound on the difference between the two invariant measures depends on the numerical approximation step size $\eta$, the inverse temperature parameter $\beta$ and the upper bound $C_{\psi}$. We emphasize that the dependence on $\beta$ is reasonable since different $\beta$ results in different diffusion, and further leads to different stationary distributions of the SDE and its numerical approximations. 

\noindent\textbf{Bounding $I_3$: Gap between Langevin Diffusion and Global Minimum}

Most existing studies \citep{welling2011bayesian,sato2014approximation,chen2015convergence} on Langevin dynamics based algorithms focus on the convergence of the averaged sample path to the stationary distribution. 
The property of Langevin diffusion asymptotically concentrating on the global minimum of $F_n$ is well understood \citep{chiang1987diffusion,gelfand1991recursive} , which makes the convergence to a global minimum possible, even when the function $F_n$ is nonconvex. 

We give an explicit bound between the stationary distribution of Langevin diffusion and the global minimizer of $F_n$, i.e., the last term $\EE[F_n(\bX^{\pi})]-F_n(\xb^*)$ in \eqref{eq:decomp_excess_risk}. For nonconvex objective function, this has been proved in \cite{raginsky2017non} using the concept of differential entropy and smoothness of $F_n$. We formally summarize it as the following lemma:
\begin{lemma}\label{thm:minimizer_Gibbs}\cite{raginsky2017non}
Under Assumptions \ref{assump:smooth} and \ref{assump:dissipative}, the model error $I_3$ in \eqref{eq:decomp_excess_risk} can be bounded by
\begin{align*}
\EE [F_n(\bX^{\pi})]-F_n(\xb^*)\leq\frac{d}{2\beta}\log\bigg(\frac{eM(m\beta/d+1)}{m}\bigg),
\end{align*}
where $\bX^{\pi}$ is a random vector following the stationary distribution of Langevin diffusion \eqref{eq:langevin diffusion}.
\end{lemma}

Lemma \ref{thm:minimizer_Gibbs} suggests that Gibbs density concentrates on the global minimizer of objective function. Therefore, the random vector $\bX^{\pi}$ following the Gibbs distribution $\pi$ is also referred to as an \textit{almost minimizer} of the nonconvex function $F_n$ in \cite{raginsky2017non}. 

\subsection{Proof of Theorems \ref{thm:converge_gld}, \ref{thm:converge_sgld} and \ref{thm:converge_vrsgld}}
 
Now we integrate the previous lemmas to prove our main theorems in Section \ref{sec:main_result}. First, submitting the results in Lemmas \ref{lemma:ergodic_gld}, \ref{theorem:density_dist} and  \ref{thm:minimizer_Gibbs} into \eqref{eq:decomp_excess_risk}, we immediately obtain the optimization error bound in \eqref{eq:err_bound_gld} for GLD, which proves Theorem \ref{thm:converge_gld}.
Second, consider the optimization error of SGLD (Algorithm \ref{alg:sgld}), 
we only need to bound the error between $\EE[F_n(\bY_K)]$ and $\EE[F_n(\bX_K)]$ and then apply the results for GLD, which is given by the following lemma.
\begin{lemma}\label{thm:ergodic_sgld}
Under Assumptions \ref{assump:smooth} and \ref{assump:dissipative}, by choosing mini-batch of size $B$,
the output of SGLD in Algorithm \ref{alg:sgld} ($\bY_K$) and the output of GLD in Algorithm \ref{alg:gld} ($\bX_K$) satisfies
\begin{align}
    |\EE[F_n(\bY_K)]-\EE[F_n(\bX_K)]|
    \leq C_1 \sqrt{\beta}\Gamma (M\sqrt{\Gamma}+G)  K\eta  \bigg[ \frac{n-B}{B(n-1)} \bigg]^{1/4},
\end{align}
where $C_1$ is an absolute constant and $\Gamma=2(1+1/m)(b+2G^2+d/\beta)$.
\end{lemma}
\noindent Combining Lemmas \ref{lemma:ergodic_gld}, \ref{theorem:density_dist},   \ref{thm:minimizer_Gibbs} and \ref{thm:ergodic_sgld} yields the desired result in \eqref{thm:converge_sgld} for SGLD, which completes the proof of Theorem \ref{thm:converge_sgld}.
Third, similar to the proof of SGLD, we require an additional bound between $F_n(\bZ_K)$ and $F_n(\bX_K)$ for the proof of SVRG-LD, which is stated by the following lemma.
\begin{lemma}\label{thm:ergodic_vrsgld}
Under Assumptions \ref{assump:smooth} and \ref{assump:dissipative}, by choosing mini-batch of size $B$,
the output of SVRG-LD in Algorithm \ref{alg:vr_sgld} ($\bZ_K$) and the output of GLD in Algorithm \ref{alg:gld} ($\bX_K$) satisfies
\begin{align*}
    &\big|\EE[F_n(\bZ_K)]-\EE[F_n(\bX_K)]\big|\leq C_1\Gamma K^{3/4} \eta \bigg[ \frac{L M^2(n-B)( 3L\eta\beta (M^2 \Gamma+G^2) +  d/2)}{B(n-1)}\bigg]^{1/4},
\end{align*}
where $\Gamma=2(1+1/m)(b+2G^2+d/\beta)$,  $C_1$ is an absolute constant and $L$ is the number of inner loops in SVRG-LD.
\end{lemma}
\noindent The optimization error bound in \eqref{eq:err_bound_vr_sgld} for SVRG-LD follows from Lemmas \ref{lemma:ergodic_gld}, \ref{theorem:density_dist},   \ref{thm:minimizer_Gibbs} and \ref{thm:ergodic_vrsgld}.


\section{Conclusions and Future Work}\label{sec:conclusion}
In this work, we present a new framework for analyzing the convergence of Langevin dynamics based algorithms, and provide non-asymptotic analysis on the convergence for nonconvex finite-sum optimization.
By comparing the Langevin dynamics based algorithms and standard first-order optimization algorithms, we may see that the counterparts of GLD and SVRG-LD are gradient descent (GD) and stochastic variance-reduced gradient (SVRG) methods. It has been proved that SVRG outperforms GD universally for nonconvex finite-sum optimization \citep{reddi2016stochastic,allen2016variance}. This poses a natural question that whether SVRG-LD can be universally better than GLD for nonconvex optimization? We will attempt to answer this question in the future.


\section*{Acknowledgement}
We would like to thank the anonymous reviewers for their helpful comments. We thank Maxim Raginsky for insightful comments and discussion on the first version of this paper. We also thank Tianhao Wang for discussion on this work. This research was sponsored in part by the National Science Foundation IIS-1652539. The views and conclusions contained in this paper are those of the authors and should not be interpreted as representing any funding agencies.

\bibliographystyle{plainnat}
\bibliography{reference}
\appendix

\section{Fokker-Planck Equation and Backward Kolmogorov Equation}
In this section, we introduce the Fokker-Planck Equation and the Backward Kolmogorov equation. 
Fokker-Planck equation addresses the evolution of probability density $p(\xb)$ that associates with the SDE. We give the following specific definition.
\begin{definition}[Fokker--Planck Equation]
Let $p(\xb,t)$ be the probability density at time $t$ of the stochastic differential equation and denote $p_0(\xb)$ the initial probability density. Then
\begin{align*}
    \partial_t p(\xb,t) = \cL^* p(\xb,t), \quad p(\xb, 0) = p_0(\xb),
\end{align*}
where $\cL^*$ is the formal adjoint of $\cL$.
\end{definition} 
Fokker-Planck equation gives us a way to find whether there exists a stationary distribution for the SDE. It can be shown \citep{ikeda2014stochastic} that for the stochastic differential equation \eqref{eq:langevin diffusion}, its stationary distribution exists and satisfies
\begin{align}\label{eq:gibbs}
     \pi(d\xb)= \frac{1}{Q}e^{-\beta F_n(\xb)}, \quad Q = \int e^{-\beta F_n(\xb)} d\xb.
\end{align}
This is also known as Gibbs measure.

Backward Kolmogorov equation describes the evolution of $\EE[g(\bX(t))|\bX(0 )= \xb]$ with $g$ being a smooth test function.

\begin{definition}[Backward Kolmogorov Equation]
Let $\bX(t)$ solves the stochastic differential equation \eqref{eq:langevin diffusion}. Let $u(\xb,t) = \EE[g(\bX(t))|\bX(0) = \xb]$, we have
\begin{align*}
    \partial_t u(\xb,t) = \cL u(\xb,t), \quad u(\xb, 0) = g(\xb).
\end{align*}
\end{definition}   
Now consider doing first order Taylor expansion on $u(\xb,t)$, we have
\begin{align}\label{eq:1j1111}
    u(\xb, t) &= u(\xb, 0) + \frac{\partial}{\partial t}u(\xb, t)|_{t=0}\cdot(t -0) + O(t^2)\notag\\
    &= g(\xb) + t\cL g(\xb) + O(t^2).
\end{align}

\section{Proof of Corollaries}
In this section, we provide the proofs of corollaries for iteration complexity in our main theory section.

\begin{proof}[Proof of Corollary \ref{coro:gld}]
To ensure the iterate error converge to $\epsilon$ precision, we need
\begin{align*}
    \Theta  e^{-\lambda K\eta}\leq\frac{\epsilon}{2},\qquad \frac{C_{\psi}\eta}{\beta}\leq\frac{\epsilon}{2}.
\end{align*}
The second inequality can be easily satisfied  with $\eta=O(\epsilon)$ and the first inequality implies
\begin{align*}
    K\geq\frac{1}{\lambda\eta}\log\bigg(\frac{2\Theta}{\epsilon}\bigg).
\end{align*}
Combining with $\eta=O(\epsilon)$ and $\Theta=O(d^2/\rho^{d/2})$, we obtain the iteration complexity 
\begin{align*}
K=O\bigg(\frac{d}{\epsilon\lambda}\cdot \log\bigg(\frac{1}{\epsilon }\bigg) \bigg),
\end{align*}
which completes the proof.
\end{proof}

\begin{proof}[Proof of Corollary \ref{coro:sgld}]
To ensure the iterate error of SGLD converging to $\epsilon$ precision, we require the following inequalities to hold
\begin{align*}
    C_1 \sqrt{\beta}\Gamma (M\sqrt{\Gamma}+G)  K\eta  \bigg[ \frac{n-B}{B(n-1)} \bigg]^{1/4} \leq \frac{\epsilon}{3}, \qquad \Theta  e^{-\lambda K\eta}\leq\frac{\epsilon}{3},\qquad \frac{C_{\psi}\eta}{\beta}\leq\frac{\epsilon}{3}.
\end{align*}
The third inequality can be easily satisfied  with $\eta=O(\epsilon)$. For the second inequality, similar as in the proof of Corollary \ref{coro:gld}, we have
\begin{align*}
        K\eta\geq\frac{1}{\lambda}\log\bigg(\frac{3\Theta}{\epsilon}\bigg).
\end{align*}
Since $\epsilon<1$, we know that $\log(1/\epsilon)$ will not go to zero when $\epsilon$ goes to zero. In fact, if we set $\eta=O(\epsilon)$ and $K=O(d/(\lambda\epsilon)\log(1/\epsilon))$, the first term in \eqref{eq:err_bound_sgld} scales as
\begin{align*}
   C_1 \sqrt{\beta}\Gamma (M\sqrt{\Gamma}+G)  K\eta  \bigg[ \frac{n-B}{B(n-1)} \bigg]^{1/4} &= O\bigg(\frac{d^{3/2} K\eta}{B^{1/4}} \bigg) = O\bigg(\frac{d^{3/2}}{B^{1/4} \lambda }\log\bigg(\frac{1}{\epsilon}\bigg)\bigg).
\end{align*}
Therefore, within $K = O(d/(\epsilon\lambda)\cdot\log(1/\epsilon))$ iterations, the iterate error of SGLD scales as
\begin{align*}
    O\bigg(\frac{d^{3/2}}{B^{1/4} \lambda }\log \bigg(\frac{1}{\epsilon}\bigg) + \epsilon\bigg).
\end{align*}
This completes the proof.
\end{proof}

\begin{proof}[Proof of Corollary \ref{coro:vr_sgld}]
Similar to previous proofs, in order to achieve an $\epsilon$-precision iterate error for SVRG-LD, we require 
\begin{align*}
    C_1\Gamma K^{3/4} \eta \bigg[ \frac{L\beta M^2(n-B)}{B(n-1)}\bigg( 9\eta (M^2 \Gamma+G^2) + \frac{ d}{\beta}\bigg)\bigg]^{1/4}\leq\frac{\epsilon}{3},\quad\Theta  e^{-\lambda K\eta}\leq\frac{\epsilon}{3},\quad \frac{C_{\psi}\eta}{\beta}\leq\frac{\epsilon}{3}.
\end{align*}
By previous proofs we know that the second and third inequalities imply $\eta=O(\epsilon)$ and $K\eta=O(1/\lambda\log(3\Theta/\epsilon))$ respectively. Combining with the first inequality, we have
\begin{align*}
    \eta^{1/4}=O\bigg( \frac{B^{1/4}\epsilon}{(K\eta)^{3/4}d^{5/4}L^{1/4}} \bigg).
\end{align*}
Combining with the first inequality, we have
\begin{align*}
    \eta=O\bigg( \min\bigg\{ \frac{B \epsilon^4}{(K\eta)^{3}d^{5}L}, \epsilon \bigg\}\bigg).
\end{align*}
Combining the above requirements yields
\begin{align}\label{eq:yyyy1}
    K=O\bigg(\frac{Ld^5}{B\lambda^4\epsilon^4}\log^4\bigg(\frac{1}{\epsilon }\bigg) + \frac{1}{\epsilon} \bigg).
\end{align}
For gradient complexity, note that for each iteration we need $B$ stochastic gradient evaluations and we also need in total $K/L$ full gradient calculations. Therefore, the gradient complexity for SVRG-LD is 
\begin{align*}
    O(K\cdot B + K/L \cdot n) =\tO\bigg(\bigg(\frac{n}{B} + L\bigg) \frac{1}{\epsilon^4}+ \bigg(\frac{n}{L} + B\bigg) \frac{1}{\epsilon}  \bigg) \cdot e^{\tO(d)}.
\end{align*}
If we solve for the best $B$ and $L$, we obtain $B =  \sqrt{n}\epsilon^{-3/2}$, $L =  \sqrt{n}\epsilon^{3/2}$. Therefore,
we have the optimal gradient complexity for SVRG-LD as
\begin{align*}
    \tO\bigg( \frac{\sqrt{n}}{\epsilon^{5/2}}   \bigg)\cdot e^{\tO(d)},
\end{align*}
which completes the proof.
\end{proof}

\section{Proof of Technical Lemmas}\label{app:tech}
In this section, we provide proofs of the technical lemmas used in the proof of our main theory.
\subsection{Proof of Lemma \ref{lemma:ergodic_gld} }\label{app:erg}
Geometric ergodicity of dynamical systems has been studied a lot in the literature  \citep{roberts1996exponential,mattingly2002ergodicity}. In particular, \citet{roberts1996exponential} proved that even when the diffusion converges exponentially fast to its stationary distribution, the Euler-Maruyama discretization in \eqref{eq:sgld} may still lose the convergence properties and examples for Langevin diffusion can be found therein. To further address this problem, \cite{mattingly2002ergodicity} built their analysis of ergodicity for SDEs on a \textit{minorization} condition and the existence of a Lyapunov function. In time discretization of dynamics systems, they studied how time-discretization affects the minorization condition and the Lyapunov structure. For the self-containedness of our analysis, we present the minorization condition on a compact set $\cC$ as follows. 
\begin{proposition}\label{prop:minor}
There exist $t_0\in \RR$ and $\xi>0$ such that the Markov process $\{\bX(t)\}$ satisfies
\begin{align*}
    \PP(\bX(t_0)\in A|\bX(0)=\xb)\geq \xi\nu(A),
\end{align*}
for any $A\in\cB(\RR^d)$, some fixed compact set $\cC\in\cB(\RR^d)$, and $\xb\in\cC$, where $\cB(\RR^d)$ denotes the Borel $\sigma$-algebra on $\RR^d$ and $\nu$ is a probability measure with $\nu(\cC^c)=0$ and $\nu(\cC)=1$.
\end{proposition}
Proposition \ref{prop:minor} does not always hold for a Markov process generated by an arbitrary SDE. However, for Langevin diffusion \eqref{eq:langevin diffusion} studied in this paper, \citet{mattingly2002ergodicity} proved that this minorization condition actually holds under the dissipative and smooth assumptions (see Corollary $7.4$ in \cite{mattingly2002ergodicity}). For more explanation on the existence and robustness of the minorization condition under discretization approximations for Langevin diffusion, we refer interested readers to Corollary $7.5$ and the proof of Theorem $6.2$ in \cite{mattingly2002ergodicity}. Now we are going to prove Lemma \ref{lemma:ergodic_gld}, which requires the following useful lemmas:
\begin{lemma}\label{lemma:Lya_bound}
Let $V(\xb)=C+\|\xb\|_2^2$ be a function on $\RR^d$, where $C>0$ is a constant. Denote the expectation with Markov process $\{\bX(t)\}$ starting at $\xb$ by $\EE^{\xb}[\cdot]=\EE[\cdot|\bX(0)=\xb]$. Under Assumption \ref{assump:dissipative}, we have
\begin{align*}
    \EE_{}^{\xb}[V(\bX(t))]\leq e^{-2mt}V(\xb)+\frac{b+m+d/\beta}{m}(1-e^{-2mt}),
\end{align*}
for all $\xb\in\RR^d$. 
\end{lemma}
\begin{lemma}(Theorem $7.3$ in \cite{mattingly2002ergodicity})\label{lemma:theorem_7_3_mattingly}
Under Assumptions \ref{assump:smooth} and \ref{assump:dissipative}, let $V(\xb)=C_0+M/2\|\xb\|_2^2$ be an essential quadratic function. The numerical approximation \eqref{eq:gld} (GLD) of Langevin diffusion \eqref{eq:langevin diffusion} has a unique invariant measure $\mu$ and for all test function $g$ such that $|g|\leq V$, we have
\begin{align*}
    \big|\EE[g(\bX_k)]-\EE[g(\bX^{\mu})]\big|\leq C\kappa\rho_{\beta}^{-d/2}(1+\kappa e^{m\eta})\exp\bigg(-\frac{2mk\eta\rho_{\beta}^d}{\log(\kappa)}\bigg),
\end{align*}
where $C>0$ is an absolute constant, $\rho_{\beta}\in(0,1)$ is some contraction parameter depending on the inverse temperature $\beta$ of Langevin dynamics, and $\kappa=2M(b+m+d)/m$.
\end{lemma}
\begin{proof}[Proof of Lemma \ref{lemma:ergodic_gld}]
The proof is majorly adapted from that of Theorem $7.3$ and Corollary $7.5$ in \cite{mattingly2002ergodicity}. By Assumption \ref{assump:smooth}, $F_n$ is $M$-smooth. Thus we have
\begin{align*}
    F_n(\xb)\leq F_n(\yb)+\la\nabla F_n(\yb),\xb-\yb\ra+\frac{M}{2}\|\xb-\yb\|_2^2,
\end{align*}
for all $\xb,\yb\in\RR^d$. By Lemma \ref{lemma:gradient_bound} and choosing $\yb = \zero$, this immediately implies that $F_n(\xb)$ can always be bounded by a quadratic function $V(\xb)$, i.e.,
\begin{align*}
    F_n(\xb)\leq\frac{M}{2}V(\xb)=\frac{M}{2}(C_0+\|\xb\|_2^2).
\end{align*}
Therefore $V(\xb)$ is an essentially quadratic Lyapunov function such that $|F_n(\xb)|\leq MV(\xb)/2$ for $\xb\in\RR^d$. By Lemma \ref{lemma:Lya_bound} the Lyapunov function satisfies
\begin{align*}
    \EE_{}^{\xb_0}[V(\bX(t))]\leq e^{-2mt}V(\xb_0)+\frac{b+m+d/\beta}{m}(1-e^{-2mt}).
\end{align*}
According to Corollary $7.5$ in \cite{mattingly2002ergodicity}, the Markov chain $\{\bX_k\}_{k=1,2,\ldots,K}$ satisfies
\begin{align}\label{eq:bound_Vx1_Vx0}
    \EE_{}^{\xb_0}[MV(\bX_1)/2]\leq e^{-2m\eta}[MV(\xb_0)/2]+\frac{M(b+m+d/\beta)}{2m}.
\end{align}
Recall the GLD update formula defined in \eqref{eq:gld}
\begin{align*} 
\bX_{k+1} = \bX_k - \eta \nabla F_n(\bX_k) + \sqrt{2\eta \beta^{-1} } \cdot\bepsilon_k.
\end{align*}
Define $F'(\bX_k) = \beta F_n(\bX_k)$ and $\eta' = \eta/\beta$, we have
\begin{align} \label{eq:rescale_gld}
\bX_{k+1} = \bX_k - \eta' \nabla F'(\bX_k) + \sqrt{2\eta'} \cdot\bepsilon_k.
\end{align}
This suggests that the result for $\beta \neq 1$ is equivalent to rescaling $\eta$ to $\eta/\beta$ and $F_n(\cdot)$ to $\beta F_n(\cdot)$. Therefore, in the following proof, we will assume that $\beta=1$ and then rescale $\eta$, $F_n(\cdot)$ at last. Similar tricks are used in \cite{raginsky2017non,zhang2017hitting}.
Under Assumptions \ref{assump:smooth} and \ref{assump:dissipative}, it is proved that Euler-Maruyama approximation of Langevin dynamics \eqref{eq:langevin diffusion} has a unique invariant measure $\mu$ on $\RR^d$. Denote $\bX^{\mu}$ as a random vector which is sampled from measure $\mu$. By Lemma \ref{lemma:theorem_7_3_mattingly}, for all test function $g$ such that $|g|\leq V$, it holds that
\begin{align*}
    \big|\EE[g(\bX_k)]-\EE[g(\bX^{\mu})]\big|\leq C\kappa'\rho_{\beta}^{-d/2}(1+\kappa' e^{m\eta})\exp\bigg(-\frac{2mk\eta\rho_{\beta}^d}{\log(\kappa')}\bigg),
\end{align*}
where $C>0$ is an absolute constant, $\rho_{\beta}\in(0,1)$ is some contraction parameter depending on the inverse temperature $\beta$ of Langevin dynamics, and $\kappa'=2M(b+m+d)/m$. Take $F_n$ as the test function and $\bX_0=\zero$, and by rescaling $\eta$ and $F_n(\cdot)$ (dissipative and smoothness parameters), we have
\begin{align*}
    \big|\EE[F_n(\bX_k)]-\EE[F_n(\bX^{\mu})]\big|\leq C\kappa\rho_{\beta}^{-d/2}(1+\kappa e^{m\eta})\exp\bigg(-\frac{2mk\eta \rho_{\beta}^d}{\log(\kappa)}\bigg),
\end{align*}
where $\kappa = 2M(b\beta+m\beta+d)/m$.
\end{proof}

\subsection{Proof of Lemma \ref{theorem:density_dist}}
To prove Lemma \ref{theorem:density_dist}, we lay down the following supporting lemma, of which the derivation  is inspired and adapted from \cite{chen2015convergence}.


\begin{lemma}\label{lemma:possion_average_beta}
Under Assumptions \ref{assump:smooth} and \ref{assump:dissipative}, the Markov chain $\{\bX_k\}_{k =1}^K$  generated by Algorithm \ref{alg:gld} satisfies
\begin{align*} 
    \bigg| \frac{1}{K}\sum_{k=0}^{K-1} \EE[F_n(\bX_k)|\bX_0 = \xb] - \bar F  \bigg| \leq C_{\psi}\bigg( \frac{\beta}{\eta K} + \frac{\eta}{\beta}\bigg),
\end{align*}
where $\bar F = \int F_n(\xb)\pi(d\xb)$ with $\pi$ being the Gibbs measure for the Langevin diffusion \eqref{eq:langevin diffusion}.
\end{lemma}


\begin{proof}[Proof of Lemma \ref{theorem:density_dist}]
By definition we have
\begin{align}\label{eq:j1111}
    \big|\EE[F_n(\bX^{\mu})]-\EE[F_n(\bX^{\pi})]\big| = \bigg| \int F_n(\xb)\mu(d\xb) - \int F_n(\xb)\pi(d\xb) \bigg|.
\end{align}
For simplicity, we denote the average $\int F_n(\xb)\pi(d\xb)$ as $\bar F_n$. Since $\mu$ is the ergodic limit of the Markov chain generated by the GLD process, for a given test function $F_n$, we have
\begin{align*}
    \int F_n(\xb)\mu(d\xb) = \int \EE[F_n(\bX_k)|\bX_0 = \xb]\cdot\mu(d\xb).
\end{align*}
Since $\mu$ and $\pi$ are two invariant measures, we consider the case where $K \rightarrow \infty$. Take average over $K$ steps $\{\bX_k\}_{k=0}^{K-1}$ we have
\begin{align}\label{eq:j2222}
    \int F_n(\xb)\mu(d\xb) = \lim_{K\rightarrow \infty} \int \frac{1}{K}\sum_{k=0}^{K-1} \EE[F_n(\bX_k)|\bX_0 = \xb]\cdot\mu(d\xb).
\end{align}
Submitting \eqref{eq:j2222} back into \eqref{eq:j1111} yields
\begin{align}\label{eq:j3333}
    \big|\EE[F_n(\bX^{\mu})]-\EE[F_n(\bX^{\pi})]\big| &=\lim_{K\rightarrow \infty} \bigg| \int \bigg[\frac{1}{K}\sum_{k=0}^{K-1} \EE[F_n(\bX_k)|\bX_0 = \xb] - \bar F \bigg]\cdot\mu(d\xb) \bigg|\notag\\
    &\leq \lim_{K\rightarrow \infty}\int\bigg| \frac{1}{K}\sum_{k=0}^{K-1} \EE[F_n(\bX_k)|\bX_0 = \xb] - \bar F  \bigg|\cdot\mu(d\xb).
\end{align}
Apply Lemma \ref{lemma:possion_average_beta} with $g$ chosen as $F_n$ we further bound \eqref{eq:j3333} by
\begin{align*} 
    \big|\EE[F_n(\bX^{\mu})]-\EE[F_n(\bX^{\pi})]\big|
    &\leq C_{\psi}\cdot\lim_{K\rightarrow \infty} \int \bigg( \frac{\beta}{\eta K} + \frac{\eta}{\beta}\bigg)   \cdot\mu(d\xb)\\
    &= C_{\psi} \cdot\lim_{K\rightarrow \infty} \bigg(\frac{\beta}{\eta K} + \frac{\eta}{\beta}\bigg)\\
    &= \frac{C_{\psi} \eta}{\beta}.
\end{align*}
This completes the proof.
\end{proof}

\subsection{Proof of Lemma \ref{thm:ergodic_sgld}}
Lemma \ref{thm:ergodic_sgld} gives the upper bound of function value gap between the GLD iterates and the SGLD iterates. To bound the difference between $F_n(\bX_K)$ and $F_n(\bY_K)$, we need the following lemmas.
\begin{lemma}\label{lemma:variance_sgd}
Under Assumptions \ref{assump:smooth} and \ref{assump:dissipative}, for any $\xb \in \RR^d$, it holds that
\begin{align*}
     \EE  \bigg\|\nabla F_n(\xb)  - \frac{1}{B}\sum_{i \in I_k} \nabla f_i(\xb) \bigg\|_2^2  \leq 
     \frac{4(n-B)(M\|\xb\|_2+G)^2}{B(n-1)}, 
\end{align*}
where $B = |I_k|$ is the mini-batch size and $G = \max_{i=1,\ldots,n}\{\|\nabla f_i(\xb^*)\|_2\} + bM/m$.
\end{lemma}
The following lemma describes the $L^2$ bound for discrete processes $\bX_k$ (GLD), $\bY_k$ (SGLD) and $\Zb_k$ (SVRG-LD). Note that for SGLD, similar result is also presented in \cite{raginsky2017non}.
\begin{lemma}\label{lemma:L2_bound_Xk}
Under Assumptions \ref{assump:smooth} and \ref{assump:dissipative}, for sufficiently small step size $\eta$, suppose the initial points of Algorithms \ref{alg:gld}, \ref{alg:sgld} and \ref{alg:vr_sgld} are chosen at $\zero$, then the $L^2$ bound of the GLD process \eqref{eq:gld}, SGLD process \eqref{eq:sgld} and SVRG-LD process \eqref{eq:SVRG-LD} can be uniformly bounded by
\begin{align*}
    \max\{\EE[\|\bX_k\|_2^2],\EE[\|\bY_k\|_2^2],\EE[\|\bZ_k\|_2^2]\}\leq \Gamma \quad\mbox{ where }\quad \Gamma := 2\bigg(1+\frac{1}{m}\bigg)\bigg(b+2G^2+\frac{d}{\beta}\bigg),
\end{align*}
for any $k=0,1,\ldots,K$, where $G = \max_{i=1,\ldots,n}\{\|\nabla f_i(\xb^*)\|_2\} + bM/m$.
\end{lemma}
The following lemma gives out the upper bound for the exponential $L^2$ bound of $\bX_k$.
\begin{lemma}\label{lemma:L2_bound_exp_Xk}
Under Assumptions \ref{assump:smooth} and \ref{assump:dissipative}, for sufficiently small step size $\eta<1$ and the inverse temperature satisfying $\beta\geq\max\{2/(m-M^2\eta),4\eta\}$, it holds that
\begin{align*}
\log \EE[\exp(\|\bX_k\|_2^2)]\le\|\bX_0\|_2^2+\frac{2\beta(b+G^2)+2d}{\beta-4\eta}k\eta.
\end{align*}
\end{lemma}

\begin{lemma}\citep{polyanskiy2016wasserstein,raginsky2017non}\label{lemma:W2_continuity}
For any two probability density functions $\mu,\nu$ with bounded second moments, let $g:\RR^d\rightarrow\RR$ be a $C^1$ function such that
\begin{align*}
\|\nabla g(\xb)\|_2\leq C_1\|\xb\|_2+C_2,  \forall \xb\in \RR^d
\end{align*}
for some constants $C_1,C_2\geq 0$. Then
\begin{align*}
\bigg|\int_{\RR^d}g(\xb)d\mu-\int_{\RR^d}g(\xb)d\nu\bigg|\leq (C_1\sigma+C_2)\cW_2(\mu,\nu),
\end{align*}
where $\cW_2$ is the $2$-Wasserstein distance and $\sigma^2=\max\big\{\int_{\RR^d}\|\xb\|_2^2\mu(d\xb),\int_{\RR^d}\|\xb\|_2^2\nu(d\xb)\big\}$.
\end{lemma}

\begin{lemma}(Corollary $2.3$ in \cite{bolley2005weighted})\label{lemma:BV_lemma}
Let $\nu$ be a probability measure on $\RR^d$. Assume that there exist $\xb_0$ and a constant $\alpha>0$ such that $\int\exp(\alpha\|\xb-\xb_0\|_2^2)d \nu(\xb)<\infty$. Then for any probability measure $\mu$ on $\RR^d$, it satisfies
\begin{align*}
    \cW_2(\mu,\nu)\leq C_{\nu}\big(\sqrt{D_{\text{KL}}(\mu||\nu)}+\big(D_{\text{KL}}(\mu||\nu)/2\big)^{1/4}\big),
\end{align*}
where $C_{\nu}$ is defined as
\begin{align*}
    C_{\nu}=\inf_{\xb_0\in\RR^d,\alpha>0}\sqrt{\frac{1}{\alpha}\bigg(\frac{3}{2}+\log\int\exp(\alpha\|\xb-\xb_0\|_2^2)d \nu(\xb)\bigg)}.
\end{align*}
\end{lemma}

\begin{proof}[Proof of Lemma \ref{thm:ergodic_sgld}]
Let $P_K, Q_K$ denote the probability measures for GLD iterate $\bX_K$ and SGLD iterate $\bY_K$ respectively. Applying Lemma \ref{lemma:W2_continuity} to probability measures $P_K$ and $Q_K$ yields
 \begin{align}\label{eq:sgld_exp_func_bound}
 \big|\EE[F_n(\bY_K)]-\EE[F_n(\bX_K)]\big|\leq (C_1\sqrt{\Gamma}+C_2)\cW_2(Q_K,P_K),
 \end{align}
where $C_1,C_2>0$ are absolute constants and
$\Gamma=2(1+1/m)(b+2G^2+d/\beta)$ is the upper bound for both $\EE[\|\bX_k\|_2^2]$ and $\EE[\|\bY_k\|_2^2]$ according to Lemma \ref{lemma:L2_bound_Xk}. We further bound the $\cW_2$ distance via the KL-divergence by Lemma \ref{lemma:BV_lemma} as follows
\begin{align}\label{eq:W2_KL_bound_sgld_sgld}
    \cW_2(Q_K,P_K)\leq \Lambda(\sqrt{D_{\text{KL}}(Q_K||P_K)}+\sqrt[4]{D_{\text{KL}}(Q_K||P_K)}),
\end{align}
where $\Lambda=\sqrt{3/2+\log\EE_{P_K}[ \exp(\|\bX_K\|_2^2)]}$. Applying Lemma \ref{lemma:L2_bound_exp_Xk} we obtain $\Lambda=\sqrt{(6+2\Gamma)K\eta}$. Therefore, we only need to bound the KL-divergence between density functions $P_K$ and $Q_K$. To this end, we introduce a continuous-time Markov process $\{\bD(t) \}_{t \geq 0}$ to bridge the gap between diffusion $\{ \bX(t) \}_{t \geq 0}$ and its numerical approximation $\{ \bX_k \}_{k = 0,1,\dots, K}$. 
Define
\begin{align}\label{eq:D_diffusion_sgld}
d\bD(t)= b(\bD(t))dt + \sqrt{2 \beta^{-1} } d \bB(t),
\end{align}
where $b(\bD(t)) = -\sum_{k = 0}^\infty \nabla F(\bX(\eta k)) \mathds{1}\{t \in \big[\eta k, \eta (k +1)\big)\} $. Integrating \eqref{eq:D_diffusion_sgld} on interval $\big[\eta k, \eta (k +1)\big)$ yields 
\begin{align*} 
\bD(\eta (k+1)) = \bD(\eta k) - \eta \nabla F(\bD(\eta k)) + \sqrt{2\eta \beta^{-1} } \cdot\bepsilon_k,
\end{align*}
where $\bepsilon_k\sim N(\zero,\Ib_{d\times d})$. This implies that the distribution of random vector $(\bX_1,\ldots,\bX_K)$ is equivalent to that of $(\bD(\eta),\ldots,\bD(\eta K))$. 
Similarly, for $\bY_k$ we define
\begin{align*}
    d\tilde\bM(t)= c(\tilde\bM(t))dt + \sqrt{2 \beta^{-1} } d \bB(t),
\end{align*}
where the drift coefficient is defined as $c(\tilde\bM(t))=-\sum_{k = 0}^\infty g_k(\tilde\bM(\eta k)) \mathds{1}\{t \in [\eta k, \eta (k +1))\}$ and $g_k(\xb)=1/B\sum_{i\in I_k}\nabla f_i(\xb)$ is a mini-batch of the full gradient with $I_k$ being a random subset of $\{1,2,\ldots,n\}$ of size $B$. Now we have that the distribution of random vector $(\bY_1, \ldots, \bY_K)$ is equivalent to that of $(\tilde\bM(\eta),\ldots,\tilde\bM(\eta K))$. However, the process $\tilde\bM(t)$ is not Markov due to the randomness of the stochastic gradient $g_{k}$. Therefore, we define the following Markov process which has the same one-time marginals \citep{gyongy1986mimicking} as
\begin{align}\label{eq:M_diffusion}
    d\bM(t)= h(\bM(t))dt + \sqrt{2 \beta^{-1} } d \bB(t),
\end{align}
where $h(\cdot)=-\EE[g_k(\tilde{\bM}(\eta k))\ind\{t \in [\eta k, \eta (k +1))\}|\tilde\bM(t)=\cdot]$ is the conditional expectation of the left end point of the interval which $\tilde\bM(t)$ lies in. Let $\PP_t$ denote the distribution of $\bD(t)$ and $\QQ_t$ denote the distribution of $\bM(t)$. By \eqref{eq:D_diffusion_sgld} and \eqref{eq:M_diffusion}, the Radon-Nikodym derivative of $\PP_t$ with respective to $\QQ_t$ is given by the following Girsanov formula \citep{liptser2013statistics}
\begin{align*}
    \frac{d \PP_t}{d \QQ_t}(\bM)&=\exp\bigg\{\sqrt{\frac{\beta}{2}}\int_{0}^{t}(h(\bM(s))-b(\bM(s)))^{\top}(d\bM(s)- h(\bM(s))ds)\\
    &\qquad-\frac{\beta}{4}\int_{0}^{t}\|h(\bM(s))-b(\bM(s))\|_2^2ds\bigg\}.
\end{align*}
Since Markov processes $\{\bD(t)\}_{t\geq 0}$ and $\{\bM(t)\}_{t\geq 0}$ are constructed based on Markov chains $\bX_k$ and $\bY_k$, by data-processing inequality the K-L divergence between $P_K$ and $Q_K$ can be bounded by
\begin{align}\label{eq:sgld_KL_bound_sgld}
    D_{KL}(Q_K||P_K)&\leq D_{KL}(\QQ_{\eta K}||\PP_{\eta K})\notag\\
    &=-\EE\bigg[ \log\bigg(\frac{d \PP_{\eta K}}{d \QQ_{\eta K}}(\bM)\bigg)\bigg]\notag\\
    &=\frac{\beta}{4}\int_{0}^{\eta K}\EE\big[\|h(\bM(r))-b(\bM(r))\|_2^2\big]d r,
\end{align}
where in the last equality we used the fact that $d\bB(t)$ follows Gaussian distribution independently for any $t\geq 0$. By definition, we know that both $h(\bM(r))$ and $b(\bM(r))$ are step functions when $r\in[\eta k,\eta(k+1))$ for any $k$. This observation directly yields
\begin{align*}
    \int_{0}^{\eta K}\EE\big[\|h(\bM(r))-b(\bM(r))\|_2^2\big]d r&\leq\sum_{k=0}^{K-1}\int_{\eta k}^{\eta (k+1)}\EE\big[\|g_k(\tilde\bM(\eta k))-\nabla F_n(\tilde\bM(\eta k))\|_2^2\big]d r\\
    &=\eta\sum_{k=0}^{K-1}\EE\big[\|g_k(\bY_{k})-\nabla F_n(\bY_{k})\|_2^2\big],
\end{align*}
where the first inequality is due to Jensen's inequality and the convexity of function $\|\cdot\|^2$, and the last equality is due to the equivalence in distribution. By Lemmas \ref{lemma:variance_sgd} and \ref{lemma:L2_bound_Xk}, we further have
\begin{align}\label{eq:sgld_int_variance_bound}
    \int_{0}^{\eta K}\EE\big[\|h(\bM(r))-b(\bM(r))\|_2^2\big]d r\leq   \frac{4\eta K(n-B)(M\Gamma+G)^2}{B(n-1)}.
\end{align}
Submitting \eqref{eq:sgld_KL_bound_sgld} and \eqref{eq:sgld_int_variance_bound} into \eqref{eq:W2_KL_bound_sgld_sgld}, we have
\begin{align}\label{eq:W2_final_bound_sgld}
    \cW_2(Q_K,P_K) &\leq \Lambda \Bigg(\sqrt{  \frac{\beta\eta K(n-B)(M\Gamma+G)^2}{B(n-1)}}+\sqrt[4]{ \frac{\beta\eta K(n-B)(M\Gamma+G)^2}{B(n-1)} } \Bigg) \notag\\
    &\leq \Lambda \sqrt{ \frac{\beta\eta K\sqrt{n-B}(M\Gamma+G)^2}{\sqrt{B(n-1)}} }.
\end{align}
Combining \eqref{eq:sgld_exp_func_bound} with \eqref{eq:W2_final_bound_sgld}, we obtain the expected function value gap between SGLD and GLD:
\begin{align*}
    |\EE[F(\bY_k)]-\EE[F(\bX_k)]|\leq C_1\Gamma\sqrt{K\eta}  \bigg[ \frac{\beta\eta K\sqrt{n-B}(M\sqrt{\Gamma}+G)^2}{\sqrt{B(n-1)}}  \bigg]^{1/2},
\end{align*}
where we adopt the fact that $K\eta > 1$ and assume that $C_1\geq C_2$.
\end{proof}

\subsection{Proof of Lemma \ref{thm:ergodic_vrsgld}}
Similar to the proof of Lemma \ref{thm:ergodic_sgld}, to bound the difference between $F_n(\bX_K)$ and $F_n(\bZ_K)$, we need the following lemmas. 
\begin{lemma}\label{lemma:variance_svrg}
Under Assumptions \ref{assump:smooth} and \ref{assump:dissipative}, for each iteration $k = sL +\ell$ in Algorithm \ref{alg:vr_sgld}, it holds that
\begin{align*}
     \EE \| \tilde\nabla_{k} -\nabla F_n(\bZ_k) \|_2^2 \leq \frac{M^2(n-B)}{B(n-1)} \EE  \big\| \bZ_k - \tilde{\bZ}^{(s)}\big\|_2^2,
\end{align*}
where $\tilde{\nabla}_{ k}= 1/B\sum_{i_k \in I_k} \big(\nabla f_{i_k}(\bZ_k)-\nabla f_{i_k}(\tilde{\bZ}^{(s)})+\nabla F_n(\tilde\bZ^{(s)}\big)$ and $B = |I_k|$ is the mini-batch size.
\end{lemma}


\begin{proof}[Proof of Lemma \ref{thm:ergodic_vrsgld}]
Denote $Q_K^{Z}$ as the probability density functions for $\bZ_K$. For the simplicity of notation, we omit the index $Z$ in the remaining part of this proof when no confusion arises. Similar as in the proof of Lemma \ref{thm:ergodic_sgld}, we first apply Lemma \ref{lemma:W2_continuity} to probability measures $P_K$ for $\bX_K$ and $Q_K^{Z}$ for $\bZ_K$, and obtain the following upper bound of function value gap
\begin{align}\label{eq:vrsgld_bound_func_value}
    |\EE[F_n(\bZ_K)]-\EE[F_n(\bX_K)]|\leq(C_1\sqrt{\Gamma}+C_2)\cW_2(Q_K^{Z}, P_K),
\end{align}
where $C_1,C_2>0$ are absolute constants and $\Gamma=2(1+1/m)(b+2G^2+d/\beta)$ is the upper bound for both $\EE[\|\bX_k\|_2^2]$ and $\EE[\|\bZ_k\|_2^2]$ according to Lemma \ref{lemma:L2_bound_Xk}.
Further by Lemma \ref{lemma:BV_lemma}, the $\cW_2$ distance can be bounded by
\begin{align}\label{eq:W2_KL_bound}
    \cW_2(Q_K^{Z},P_K)\leq \Lambda(\sqrt{D_{\text{KL}}(Q_K^{Z}||P_K)}+\sqrt[4]{D_{\text{KL}}(Q_K^{Z}||P_K)}),
\end{align}
where $\Lambda=\sqrt{3/2+\log\EE_{P_K}[ e^{\|\bX_K\|_2^2}]}$. Applying Lemma \ref{lemma:L2_bound_exp_Xk} we obtain $\Lambda=\sqrt{(6+2\Gamma)K\eta}$. Therefore, we need to bound the KL-divergence between density functions $P_K$ and $Q_K^{Z}$. 
Similar to the proof of Lemma \ref{thm:ergodic_sgld}, we define a continuous-time Markov process associated with $\bZ_k$ as follows
\begin{align*}
    d\tilde\bN(t)=p(\tilde\bN(t))dt+\sqrt{2\beta^{-1}}d\bB(t),
\end{align*}
where $p(\tilde\bN(t))=-\sum_{k=0}^{\infty}\tilde\nabla_k\ind\{t\in[\eta k,\eta(k+1))\}$ and $\tilde\nabla_k$ is the semi-stochastic gradient at $k$-th iteration of SVRG-LD. We have that the distribution of random vector $(\bZ_1,\ldots,\bZ_K)$ is equivalent to that of $(\tilde\bN(\eta),\ldots,\tilde\bN(\eta K))$. However, $\tilde\bN(t)$ is not Markov due to the randomness of $\tilde\nabla_k$. We define the following Markov process which has the same one-time marginals as $\tilde\bN(t)$
\begin{align}\label{eq:N_diffusion}
    d\bN(t)=q(\bN(t))dt+\sqrt{2\beta^{-1}}d\bB(t),
\end{align}
where $q(\cdot)=-\EE[\tilde\nabla_k\ind\{t\in[\eta k,\eta(k+1))\}|p(\tilde{\bN}(t))=\cdot]$. Let $\QQ_t^{Z}$ denote the distribution of $\bN(t)$.
By \eqref{eq:D_diffusion_sgld} and \eqref{eq:N_diffusion}, the Radon-Nikodym derivative of $\PP_t$ with respective to $\QQ_t^{Z}$ is given by the Girsanov formula \citep{liptser2013statistics}
\begin{align*}
    \frac{d \PP_t}{d \QQ_t^{Z}}(\bN)&=\exp\bigg\{\sqrt{\frac{\beta}{2}}\int_{0}^{t}(q(\bN(r))-b(\bN(r)))^{\top}(d\bN(r)- h(\bN(r))d r)\\
    &\qquad-\frac{\beta}{4}\int_{0}^{t}\|q(\bN(r))-b(\bN(r))\|_2^2d r\bigg\}.
\end{align*}
Since Markov processes $\{\bD(t)\}_{t\geq 0}$ and $\{\bN(t)\}_{t\geq 0}$ are constructed based on $\bX_k$ and $\bZ_k$, by data-processing inequality the K-L divergence between $P_K$ and $Q_K^{Z}$ in \eqref{eq:W2_KL_bound} can be bounded by
\begin{align}\label{eq:sgld_KL_bound}
    D_{\text{KL}}(Q_K^{Z}||P_K)&\leq D_{\text{KL}}(\QQ_{\eta K}^{Z}||\PP_{\eta K})\notag\\
    &=-\EE\bigg[ \log\bigg(\frac{d \PP_{\eta K}}{d \QQ_{\eta K}^{Z}}(\bN)\bigg)\bigg]\notag\\
    &=\frac{\beta}{4}\int_{0}^{\eta K}\EE\big[\|q(\bN(r))-b(\bN(r))\|_2^2\big]d r.
\end{align}
where in the last equality we used the fact that $d\bB(t)$ follows Gaussian distribution independently for any $t\geq 0$. By definition, we know that both $q(\bN(r))$ and $b(\bN(r))$ are step functions when $r\in[\eta k,\eta(k+1))$ for any $k$. 
This observation directly yields
\begin{align*}
    \int_{0}^{\eta K}\EE\big[\|q(\bN(r))-b(\bN(r))\|_2^2\big]d r&\leq\sum_{k=0}^{K-1}\int_{\eta k}^{\eta (k+1)}\EE\big[\tilde\nabla_k(\tilde\bN(\eta k))-\nabla F_n(\tilde\bN(\eta k))\|_2^2\big]d r\\
    &=\eta\sum_{k=0}^{K-1}\EE\big[\|\tilde\nabla_k(\bZ_{k})-\nabla F_n(\bZ_{k})\|_2^2\big],
\end{align*}
where the first inequality is due to Jensen's inequality and the convexity of function $\|\cdot\|_2^2$, and the last equality is due to the equivalence in distribution. Combine the above results we obtain
\begin{align}\label{eq:vrsgld_bound_KL}
    D_{\text{KL}}(Q_K^{Z}||P_K) &\leq\frac{\beta\eta}{4}\sum_{k=0}^{K-1}\EE[\|\tilde\nabla_k-\nabla F_n(\bZ_k)\|_2^2]\notag\\
    &\leq \frac{\beta\eta}{4} \sum_{s=0}^{K/L}\sum_{\ell=0}^{L-1}\EE[\|\tilde\nabla_{sL+\ell}-\nabla F_n(\bZ_{sL+\ell})\|_2^2],
\end{align}
where the second inequality follows the fact that $k = sL + \ell \leq (s+1)L$ for some $\ell=0,1,\ldots,L-1$.
Applying Lemma \ref{lemma:variance_svrg}, the inner summation in \eqref{eq:vrsgld_bound_KL} yields
\begin{align}\label{eq:vrsgld_bound_variance}
    \sum_{\ell=0}^{L-1}\EE[\|\tilde\nabla_{sL+\ell}-\nabla F_n(\bZ_{sL+\ell})\|_2^2]\leq \sum_{\ell=0}^{L-1} \frac{M^2(n-B)}{B(n-1)} \EE  \big\| \bZ_{sL+\ell} - \tilde{\bZ}^{(s)}\big\|_2^2.
\end{align}
Note that we have 
\begin{align}\label{eq:j5555}
     &\EE  \big\| \bZ_{sL+\ell} - \tilde{\bZ}^{(s)}\big\|_2^2 \notag\\
     &= \EE \bigg\| \sum_{u = 0}^{\ell-1} \eta \big( \nabla f_{i_{sL+u}}(\bZ_{sL+u})-\nabla f_{i_{sL+u}}(\tilde{\bZ}^{(s)})+\nabla F_n(\tilde{\bZ}^{(s)})\big) -\sum_{u =0}^{\ell-1} \sqrt{\frac{2\eta}{\beta}}\epsilon_{sL+\ell}  \bigg\|_2^2\notag\\
     &\leq \ell\sum_{u = 0}^{\ell-1} \EE \big[ 2\eta^2 \big\| \nabla f_{i_{sL+u}}(\bZ_{sL+u})-\nabla f_{i_{sL+u}}(\tilde{\bZ}^{(s)})+\nabla F_n(\tilde{\bZ}^{(s)}) \big\|_2^2 \big] + \sum_{u = 0}^{\ell-1}\frac{4\eta d}{\beta} \notag\\
      &\leq 4\ell\eta\bigg( 9\ell\eta (M^2 \Gamma^2+G^2) + \frac{d}{\beta}\bigg),
\end{align}
where the first inequality holds due to the triangle inequality for the first summation term, the second one follows from Lemma \ref{lemma:gradient_bound} and Lemma \ref{lemma:L2_bound_Xk}. Submit \eqref{eq:j5555} back into \eqref{eq:vrsgld_bound_variance} we have
\begin{align}\label{eq:ssss1}
    \sum_{\ell=0}^{L-1}\EE[\|\tilde\nabla_{sL+\ell}-\nabla F_n(\bZ_{sL+\ell})\|_2^2] &\leq\frac{4\eta M^2(n-B)}{B(n-1)}\sum_{\ell=0}^{L-1} \bigg( 9\ell^2\eta (M^2 \Gamma^2+G^2) + \frac{\ell d}{\beta}\bigg)\notag\\
    &\leq  \frac{4\eta M^2(n-B)}{B(n-1)}\bigg( 3L^3\eta (M^2 \Gamma+G^2) + \frac{dL^2}{2\beta}\bigg),  
\end{align}
Since \eqref{eq:ssss1} does not depend on the outer loop index $i$, submitting it into \eqref{eq:vrsgld_bound_KL} yields
\begin{align}\label{eq:ssss2}
    \frac{\beta\eta}{4}\sum_{k=0}^{K-1}\EE[\|\tilde\nabla_{k}-\nabla F_n(\bZ_{k})\|_2^2]  &\leq  \frac{\eta^2KL M^2(n-B)( 3L\eta\beta (M^2 \Gamma+G^2) +  d/2)}{B(n-1)}. 
\end{align}
Combining \eqref{eq:vrsgld_bound_func_value}, \eqref{eq:W2_KL_bound} \eqref{eq:vrsgld_bound_KL} and \eqref{eq:ssss2}, we obtain
\begin{align*}
    \big|\EE[F_n(\bZ_K)]-\EE[F_n(\bX_K)]\big|\leq
    C_1\Gamma\sqrt{K\eta} \bigg[ \frac{\eta^2KL M^2(n-B)( 3L\eta\beta (M^2 \Gamma+G^2) +  d/2)}{B(n-1)}\bigg]^{1/4}.
\end{align*}
where we use the fact that $K\eta>1$, $\eta<1$ and assume that $C_1\geq C_2$.
\end{proof}

\section{Proof of Auxiliary Lemmas}\label{sec:proof_of_aux_lemma}
In this section, we prove additional lemmas used in Appendix \ref{app:tech}.
\subsection{Proof of Lemma \ref{lemma:Lya_bound}}
\begin{proof}
Applying It\^{o}'s Lemma yields
\begin{align}
    dV(\bX(t))=-2\la \bX(t),\nabla F_n(\bX(t))\ra dt+\frac{2d}{\beta}dt+2\sqrt{\frac{2}{\beta}}\la\bX(t),d\bB(t)\ra.
\end{align}
Multiplying $e^{2m t}$ to both sides of the above equation, where $m>0$ is the dissipative constant, we obtain
\begin{align*}
    2m e^{2mt}V(\bX(t)) dt+e^{2mt}dV(\bX(t))&=2m e^{2mt}V(\bX(t)) dt-2e^{2mt}\la \bX(t),\nabla F_n(\bX(t))\ra dt\\
    &\qquad+\frac{2d}{\beta}e^{2mt}dt+\sqrt{\frac{8}{\beta}}e^{2mt}\la\bX(t),d\bB(t)\ra.
\end{align*}
We integrate the above equation from time $0$ to $t$ and have
\begin{align}\label{eq:int_Yt}
    V(\bX(t))&=e^{-2mt}V(\bX_0)+2m\int_{0}^{t} e^{2m (s-t)}V(\bX(s)) ds-2\int_{0}^{t}e^{2m(s- t)}\la \bX(s),\nabla F_n(\bX(s))\ra ds\notag\\
    &\qquad+\frac{2d}{\beta}\int_{0}^{t}e^{2m(s- t)}ds+2\sqrt{\frac{2}{\beta}}\int_{0}^{t}e^{2m(s- t)}\la\bX(s),d\bB(s)\ra.
\end{align}
Note that by Assumption \ref{assump:dissipative}, we have
\begin{align}\label{eq:dissipative_bound}
-2\int_{0}^{t}e^{2m(s- t)}\la \bX(s),\nabla F_n(\bX(s))\ra ds&\leq-2\int_{0}^{t}e^{2m(s- t)}\big(m\|\bX(s)\|_2^2-b\big) ds\notag\\
&=-2m\int_{0}^{t}e^{2m(s- t)}V(\bX(s)) ds+\frac{b+m}{m}(1-e^{-2mt}).
\end{align}
Combining \eqref{eq:int_Yt} and \eqref{eq:dissipative_bound}, and taking expectation over $\bX(t)$ with initial point $\xb$, we get
\begin{align*}
    \EE_{}^{\xb}[V(\bX(t))]&\leq e^{-2mt}V(\xb)+\frac{b+m}{m}(1-e^{-2mt})+\frac{d}{m\beta}(1-e^{-2mt})\\
    &=e^{-2mt}V(\xb)+\frac{b+m+d/\beta}{m}(1-e^{-2mt}),
\end{align*}
where we employed the fact that $d\bB(s)$ follows Gaussian distribution with zero mean and is independent with $\bX(s)$. 
\end{proof}

\subsection{Proof of Lemma \ref{lemma:theorem_7_3_mattingly}}
Here we provide a sketch of proof to refine the parameters in the results by \citet{mattingly2002ergodicity}. For detailed proof, we refer interested readers to Theorem $7.3$ in \cite{mattingly2002ergodicity}.
\begin{proof}
Denote $\kappa=2M(b+m+d)/m$ according to Lemma \ref{lemma:Lya_bound} where $b,m$ are the dissipative parameters. Following the result in \cite{mattingly2002ergodicity}, we define $\phi=\rho_{\beta}^d$ with some parameter $0<\rho_{\beta}<1$ that can depend on the inverse temperature parameter $\beta$ of the dynamics \eqref{eq:langevin diffusion}. $\phi$ is some lower bound of the minorization condition similar to that in Proposition \ref{prop:minor} but is established for the Markov chain for the discretized algorithm. Note that we actually have $\beta=1$ in \cite{mattingly2002ergodicity}. For the ease of presentation, we will omit the subscript $\beta$ when no confusion arises. Let $\{\bX_{l\tau}\}_{l=0,1,\ldots}$ be a sub-sampled chain from $\{\bX_k\}_{k=0,1,\ldots}$ at sample rate $\tau>0$. By the proof of Theorem $2.5$ in \cite{mattingly2002ergodicity}, we obtain the following result
\begin{align}\label{eq:Xk_Xmu_decomp}
    \big|\EE[g(\bX_{l\tau})]-\EE[g(\bX^{\mu})]\big|\leq \kappa[\bar{V}+1](1-\phi)^{\alpha l\tau}+\sqrt{2}V(\xb_0)\delta^{l\tau}\kappa^{\alpha l\tau/2}\frac{1}{\sqrt{\phi}},
\end{align}
where $\bX^{\mu}$ follows the invariant distribution of Markov process $\{\bX_k\}_{k=0,1,\ldots}$, $\bar V=2\sup_{\xb\in \cC}V(\xb)$ is a bounded constant, $\delta\in(e^{-2m\eta},1)$ is a constant, and $\alpha\in(0,1)$ is chosen small enough such that $\delta\kappa^{\alpha/2}\leq 1$. In particular, we choose $\alpha\in(0,1)$ such that $\delta\kappa^{\alpha/2}\leq(1-\phi)^{\alpha}$, which yields
\begin{align*}
    \alpha\leq\frac{\log(1/\delta)}{\log(\sqrt{\kappa}/(1-\phi))}\leq\frac{\log(1/\delta)}{\log(\sqrt{\kappa})},
\end{align*}
where the last inequality is due to $1-\phi< 1$. Submitting the choice of $\alpha$ into \eqref{eq:Xk_Xmu_decomp} we have
\begin{align}\label{eq:combined_exponential}
    \big|\EE[g(\bX_{l\tau})]-\EE[g(\bX^{\mu})]\big|&\leq \frac{2\sqrt{2}\kappa}{\sqrt{\phi}}[\bar V+1]V(\xb_0)(1-\phi)^{l\tau\log(1/\delta)/\log(\sqrt{\kappa})}\notag\\
    &=\frac{2\sqrt{2}\kappa}{\sqrt{\phi}}[\bar V+1]V(\xb_0)e^{l\tau\log(r)},
\end{align}
where $r=(1-\phi)^{\log(1/\delta)/\log(\sqrt{\kappa})}$ is defined as the contraction parameter. Note that by Taylor's expansion we have
\begin{align}\label{eq:TaylorExp_log_r}
    \log r=\log(1-(1-r))=-(1-r)-\frac{(1-r)^2}{2}-\frac{(1-r)^3}{3}-\ldots\leq -(1-r),
\end{align}
when $|1-r|\leq 1$. By definition $r=(1-\phi)^{\log(1/\delta)/\log(\sqrt{\kappa})}$ and $\phi=\rho^d$ where $\rho\in(0,1)$ is a constant. Since it is more interesting to deal with the situation where dimension parameter $d$ is large enough and not negligible, we can always assume that $|\phi|=\rho^d$ is sufficiently small such that for any $0<\zeta<1$
\begin{align}\label{eq:TaylorExp_r}
    (1-\phi)^\zeta=1-\zeta\phi+\zeta(\zeta-1)/2\phi^2+\ldots+\binom{\zeta}{n}(-\phi)^n+\ldots\leq1-\zeta\phi
\end{align}
by Taylor's expansion. Submitting \eqref{eq:TaylorExp_log_r} and \eqref{eq:TaylorExp_r} into \eqref{eq:combined_exponential} yields
\begin{align}
    \big|\EE[g(\bX_{l\tau})]-\EE[g(\bX^{\mu})]\big|&\leq\frac{2\sqrt{2}\kappa}{\sqrt{\phi}}[\bar V+1]V(\xb_0)\exp\bigg(-\frac{2ml\tau\eta \rho^d}{\log(\kappa)}\bigg),
\end{align}
where we chose $\delta=e^{-m\eta}$. Next we need to prove that the unsampled chain is also exponential ergodic. Let $k=l\tau+j$ with $j=0,1,\ldots,\tau-1$. We immediately get
\begin{align*}
    \big|\EE[g(\bX_{l\tau+j})]-\EE[g(\bX^{\mu})]\big|&\leq\frac{2\sqrt{2}\kappa}{\sqrt{\phi}}[\bar V+1]\EE[V(\bX_j)]\exp\bigg(-\frac{2ml\tau\eta \rho^d}{\log(\kappa)}\bigg).
\end{align*}
Since the GLD approximation \eqref{eq:gld} of Langevin is ergodic when sampled at rate $\tau=1$, we have $k=l\tau=l$ and $j=0$. Note that by Lemma A.$2$ in \cite{mattingly2002ergodicity}, we have $\cC=\{\xb:V(\xb)\leq\kappa/e^{-m\eta}\}$, which implies that $\bar V=\kappa e^{m\eta}$. Thus we obtain
\begin{align*}
    \big|\EE[g(\bX_{k})]-\EE[g(\bX^{\mu})]\big|\leq C\kappa\rho^{-d/2}(\kappa e^{m\eta}+1)\exp\bigg(-\frac{2mk\eta \rho^d}{\log(\kappa)}\bigg),
\end{align*}
where we used the fact that $\xb_0=\zero$ and $C>0$ is an absolute constant. 
\end{proof}

\subsection{Proof of Lemma \ref{lemma:possion_average_beta}}
To prove Lemma \ref{lemma:possion_average_beta}, we choose the test function in Poisson equation \eqref{eq:poisson_equation} as $g=F_n$. Given the Poisson equation, suppose we choose $g$ as $F_n$, the distance between the time average of the GLD process and the expectation of $F_n$ over the Gibbs measure can be expressed by
\begin{align}\label{eq:1j2222}
    \frac{1}{K}\sum_{k=1}^K  F_n(\bX_k) - \bar F = \frac{1}{K}\sum_{k=1}^K \cL \psi(\bX_k).
\end{align}
Note that by \cite{mattingly2010convergence,vollmer2016exploration}, we know the Poisson equation \eqref{eq:poisson_equation} defined by the generator of Langevin dynamics has a unique solution $\psi$ under Assumptions \ref{assump:smooth} and \ref{assump:dissipative}. According to Theorem 3.2 in \cite{erdogdu2018global}, the $p$-th order derivatives of $\psi$ can be bounded by some polynomial growth function with sophisticated coefficients ($p=0,1,2$). To simplify the presentation, we hence follow the convention in the line of literature \citep{chen2015convergence,vollmer2016exploration} and assume that $\EE\big[\|\nabla^p\psi(\bX_k)\|\big]$ can be further upper bounded by a constant $C_{\psi}$ for all $\{\bX_k\}_{k\ge 0}$ and $p = (0,1,2)$, which is determined by the Langevin diffusion and its Poisson equation. In fact, \citet{erdogdu2018global} showed that the upper bound of derivatives (up to fourth order) of $\psi$ only requires the dissipative and smooth assumptions. We refer interested readers to \cite{erdogdu2018global} for more details on deriving the $C_{\psi}$ for Langevin diffusion. We show that the case $p=0$ can be easily verified as follows. By Assumption \ref{assump:smooth}, using a similar argument as in the proof of Lemma \ref{lemma:ergodic_gld}, we bound $F_n(\xb)$ by a quadratic function $V(\xb)$
\begin{align*}
    F_n(\xb)\leq\frac{M}{2}V(\xb)=\frac{M}{2}(C_0+\|\xb\|_2^2).
\end{align*}
Applying Assumption \ref{assump:dissipative} and Theorem $13$ in \cite{vollmer2016exploration} we have
\begin{align}\label{eq:t1111}
    |\psi(\xb)| \leq C_1(1+\|\xb\|_2^2) \leq C_2 V(\xb).
\end{align}
Note that by Assumptions \ref{assump:smooth} and \ref{assump:dissipative} we can verify that a quadratic $V(\xb)$ and $p^* = 2$ satisfy Assumption $12$ in  \cite{vollmer2016exploration} and therefore we obtain that for all $p \leq p^*$, we have
\begin{align}\label{eq:t2222}
    \sup_k \EE V^p(\bX_k) \leq \infty.
\end{align}
Combining \eqref{eq:t1111} and \eqref{eq:t2222} we show that $\psi(\bX_k)$ is bounded in expectation.

\begin{proof}
For the simplicity of notation, we first assume that $\beta = 1$ and then show the result for arbitrary $\beta$ by a scaling technique. Note that for the continuous-time Markov process $\{\bD(t)\}_{t\geq 0}$ defined in \eqref{eq:D_diffusion_sgld}, the distribution of random vector $(\bX_1,\ldots,\bX_K)$ is equivalent to that of $(\bD(\eta),\ldots,\bD(\eta K))$. Let $\psi$ be the solution of Poisson equation $\cL\psi=g-\int g(\xb)\pi(d\xb)$. Since we have $\EE[\psi(\bX_k)|\bX_0 = \xb] = \EE[\psi(\bD(\eta k))|\bD_0 = \xb]$. We denote $\EE[\psi(\bD(\eta k))|\bD_0 = \xb]$ by $\EE^{\xb}[\psi(\bD(\eta k))]$. By applying \eqref{eq:1j1111}, we compute the Taylor expansion of  $\EE^{\xb}[\psi(\bD(\eta k))]$ at $\bD(\eta (k-1))$:
\begin{align*}
    \EE^{\xb}[\psi(\bD(\eta k))] = \EE^{\xb}[\psi(\bD(\eta (k-1)))] + \eta \EE^{\xb}[\cL \psi(\bD(\eta (k-1)))] + O( \eta^2).
\end{align*}
Note that the remainder also depends on the second order derivative of the Poisson equation and are bounded by constant $C_{\psi}$. Take average over $k =1, \dots, K$ and rearrange the equation we have
\begin{align}\label{eq:1j3333}
    \frac{1}{\eta K} \big( \EE^{\xb}[\psi(\bD(\eta K))] -\psi(\xb) \big)+O( \eta) &= \frac{1}{K}\sum_{k=1}^K  \EE^{\xb}[\cL \psi(\bD(\eta (k-1))) ].
\end{align}
Submit the Poisson equation \eqref{eq:1j2222} into the above equation \eqref{eq:1j3333} we have
\begin{align*} 
     \frac{1}{K}\sum_{k=0}^{K-1}  \EE^{\xb}[F_n(\bX_{k}) ] - \bar F = \frac{1}{K}\sum_{k=1}^K  \EE^{\xb}[\cL \psi(\bX_{k-1}) ] &= \frac{1}{K}\sum_{k=1}^K  \EE^{\xb}[\cL \psi(\bD(\eta (k-1))) ]\\
     &= \frac{1}{\eta K} \big( \EE^{\xb}[\psi(\bD(\eta K)) ] -\psi(\xb) \big) + O(\eta)\\
     &= \frac{1}{\eta K} \big( \EE^{\xb}[\psi(\bX_K) ] -\psi(\xb) \big) + O(\eta),
\end{align*}
where the second and the fourth equation hold due to the fact that the distribution of $\{\bX_k\}$ is the same as the distribution of $\{\bD(\eta k)\}$. We have assumed that $\psi(\bX_k)$ and its first and second order derivatives are bounded by constant $C_{\psi}$ in expectation over the randomness of $\bX_k$. Therefore, we are able to obtain the following conclusion
\begin{align*}
      \bigg| \frac{1}{K}\sum_{k=0}^{K-1}  \EE^{\xb}[F_n(\bX_{k}) ] - \bar F \bigg| &\leq C_{\psi}\bigg( \frac{1}{\eta K} + \eta \bigg).
\end{align*} 
This completes the proof for the case $\beta = 1$. In order to apply our analysis to the case where $\beta$ can take any arbitrary constant value, we conduct the same scaling argument as in \eqref{eq:rescale_gld}.
\begin{align*} 
     \bigg| \frac{1}{K}\sum_{k=0}^{K-1}  \EE^{\xb}[F_n(\bX_{k}) ] - \bar F \bigg| \leq C_{\psi}\bigg( \frac{1}{\eta' K} + \eta'\bigg) = C_{\psi}\bigg( \frac{\beta}{\eta K} + \frac{\eta}{\beta}\bigg).
\end{align*}
This completes the proof.
\end{proof}

\subsection{Proof of Lemma \ref{lemma:variance_sgd}}
We first lay down the following lemma on the bounds of gradient of $f_i$.
\begin{lemma}\label{lemma:gradient_bound}
For any $\xb\in\RR^d$, it holds that $$\|\nabla f_i(\xb)\|_2 \leq M\|\xb\|_2+G$$ for constant $G = \max_{i=1,\ldots,n}\{\|\nabla f_i(\xb^*)\|_2\} + bM/m$.
\end{lemma}

\begin{proof}[Proof of Lemma \ref{lemma:variance_sgd}]
Let $\ub_i(\xb) = \nabla F(\xb) - \nabla f_i(\xb)$, consider
\begin{align}\label{eq:3j1111}
    \EE \bigg\| \frac{1}{B}\sum_{i \in I_k} \ub_i(\xb) \bigg\|_2^2 &= \frac{1}{B^2} \EE \sum_{i \neq i' \in I_k} \ub_i(\xb)^\top \ub_{i'}(\xb) + \frac{1}{B}\EE\|\ub_i(\xb)\|_2^2\notag\\
    &= \frac{B-1}{Bn(n-1)} \sum_{i \neq i'} \ub_i(\xb)^\top \ub_{i'}(\xb) + \frac{1}{B}\EE\|\ub_i(\xb)\|_2^2\notag\\
    &= \frac{B-1}{Bn(n-1)} \sum_{i, i'} \ub_i(\xb)^\top \ub_{i'}(\xb) -\frac{B-1}{B(n-1)}\EE\|\ub_i(\xb)\|_2^2 + \frac{1}{B}\EE\|\ub_i(\xb)\|_2^2\notag\\
    &= \frac{n-B}{B(n-1)}\EE\|\ub_i(\xb)\|_2^2,
\end{align}
where the last equality is due to the fact that $1/n\sum_{i=1}^n \ub_i(\xb) = 0$. By Lemma \ref{lemma:gradient_bound} we have $\|\nabla f_i(\xb)\|_2 \leq M\|\xb\|_2+G$, therefore we have $\|\nabla F(\xb)\|_2 \leq M\|\xb\|_2+G$ and consequently,
$\|\ub_i(\xb)\|_2 \leq 2(M\|\xb\|_2+G)$. Thus \eqref{eq:3j1111} can be further bounded as:
\begin{align*}
    \EE \bigg\| \frac{1}{B}\sum_{i \in I_k} \ub_i(\xb) \bigg\|_2^2  \leq \frac{n-B}{B(n-1)} 4(M\|\xb\|_2+G)^2.
\end{align*}
This completes the proof.
\end{proof}

\subsection{Proof of Lemma \ref{lemma:L2_bound_Xk}}
In this section, we provide the proof of $L^2$ bound of GLD and SVRG-LD iterates $\bX_k$ and $\bZ_k$. Note that a similar result of SGLD has been proved by \citet{raginsky2017non} and thus we omit the corresponding proof for the simplicity of presentation. 
\begin{proof}[Proof of Lemma \ref{lemma:L2_bound_Xk}]
\textbf{Part I}: We first prove the the upper bound for GLD. By the definition in \eqref{eq:gld}, we have
\begin{align*}
    \EE[\|\bX_{k+1}\|_2^2]&=\EE[\|\bX_k-\eta\nabla F_n(\bX_k)\|_2^2]+\sqrt{\frac{8\eta}{\beta}}\EE[\langle\bX_k-\eta\nabla F_n(\bX_k),\bepsilon_k\rangle]+\frac{2\eta}{\beta}\EE[\|\bepsilon_k\|_2^2]\\
    &=\EE[\|\bX_k-\eta\nabla F_n(\bX_k)\|_2^2]+\frac{2\eta d}{\beta},
\end{align*}
where the second equality follows from that $\bepsilon_k$ is independent on $\bX_k$. Now we bound the first term 
\begin{align*}
\EE[\|\bX_k-\eta\nabla F_n(\bX_k)\|_2^2]&=\EE[\|\bX_k\|_2^2]-2\eta\EE[\langle\bX_k,\nabla F_n(\bX_k)\rangle]+\eta^2\EE[\|\nabla F_n(\bX_k)\|_2^2]\\
&\leq \EE[\|\bX_k\|_2^2]+2\eta(b-m\EE[\|\bX_k\|_2^2])+2\eta^2(M^2\EE[\|\bX_k\|_2^2]+G^2)\\
&=(1-2\eta m+2\eta^2M^2)\EE[\|\bX_k\|_2^2]+2\eta b+2\eta^2G^2,
\end{align*}
where the inequality follows from Assumption \ref{assump:dissipative}, Lemma \ref{lemma:gradient_bound} and triangle inequality. Substitute the above bound back and we will have 
\begin{align}\label{eq:4.8.1}
\EE[\|\bX_{k+1}\|_2^2]\leq(1-2\eta m+2\eta^2M^2)\EE[\|\bX_k\|_2^2]+2\eta b+2\eta^2G^2+\frac{2\eta d}{\beta}.
\end{align}
For sufficient small $\eta$ that satisfies $\eta \leq \min\big\{1,m/(2M^2)\big\}$,
there are only two cases we need to take into account:\\
If $1-2\eta m+2\eta^2M^2\leq 0$, then from \eqref{eq:4.8.1} we have
\begin{align}\label{eq:4.8.2}
    \EE[\|\bX_{k+1}\|_2^2]&\leq 2\eta b+2\eta^2G^2+\frac{2\eta d}{\beta}\leq \|\bX_0\|_2^2+2\bigg(b+G^2+\frac{d}{\beta}\bigg).
\end{align}
If $0<1-2\eta m+2\eta^2M^2\leq 1$, then iterate \eqref{eq:4.8.1} and we have
\begin{align}\label{eq:4.8.3}
    \EE[\|\bX_{k}\|_2^2]&\leq (1-2\eta m+2\eta^2M^2)^k\|\bX_0\|_2^2+\frac{\eta b+\eta^2G^2+\frac{\eta d}{\beta}}{\eta m-\eta^2M^2}\leq \|\bX_0\|_2^2+\frac{2}{m}\bigg(b+G^2+\frac{d}{\beta}\bigg).
\end{align}
Combine \eqref{eq:4.8.2} and \eqref{eq:4.8.3} and we have
\begin{align*}
    \EE[\|\bX_{k}\|_2^2]&\leq\|\bX_0\|_2^2+\bigg(2+\frac{2}{m}\bigg)\bigg(b+G^2+\frac{d}{\beta}\bigg)= 2\bigg(1+\frac{1}{m}\bigg)\bigg(b+G^2+\frac{d}{\beta}\bigg),
\end{align*}
where the equation holds by choosing $\bX_0 = \zero$.

\noindent\textbf{Part II}: Now we prove the $L^2$ bound for SVRG-LD, i.e., $\EE[\|\bZ_{k}\|_2^2]$, by mathematical induction. Since $\tilde{\nabla}_{k}=1/B\sum_{i_k \in I_k} \big( \nabla f_{i_k}(\bZ_k)-\nabla f_{i_k}(\tilde{\bZ}^{(s)})+\nabla F_n(\tilde{\bZ}^{(s)}) \big)$, we have
\begin{align}\label{eq:2uuuu1}
    \EE[\|\bZ_{k+1}\|_2^2]&=\EE[\|\bZ_k-\eta\tilde{\nabla}_{k}\|_2^2]+\sqrt{\frac{8\eta}{\beta}}\EE[\langle\bZ_k-\eta\tilde{\nabla}_{k},\bepsilon_k\rangle]+\frac{2\eta}{\beta}\EE[\|\bepsilon_k\|_2^2]\notag\\
    &=\EE[\|\bZ_k-\eta\tilde{\nabla}_{k}\|_2^2]+\frac{2\eta d}{\beta},
\end{align}
where the second equality follows from the fact that $\bepsilon_k$ is independent of $\bZ_k$ and standard Gaussian. We prove it by induction. First, consider the case when $k =1$. Since we choose the initial point at $\bZ_0 = \zero$, we immediately have
\begin{align*} 
    \EE[\|\bZ_{1}\|_2^2]&=\EE[\|\bZ_0-\eta\tilde{\nabla}_{0}\|_2^2]+\sqrt{\frac{8\eta}{\beta}}\EE[\langle\bZ_0-\eta\tilde{\nabla}_{0},\bepsilon_0\rangle]+\frac{2\eta}{\beta}\EE[\|\bepsilon_0\|_2^2]\notag\\
    &=\eta^2 \EE[\|\nabla F_n(\bZ_0)\|_2^2] +\frac{2\eta d}{\beta}\\
    &\leq \eta^2 G^2 +\frac{2\eta d}{\beta},
\end{align*}
where the second equality holds due to the fact that $\tilde{\nabla}_{0} = \nabla F_n(\bZ_0)$ and the inequality follows from Lemma \ref{lemma:gradient_bound}. For sufficiently small $\eta$ we can see that the conclusion of Lemma \ref{lemma:L2_bound_Xk} holds for $\EE[\|\bZ_{1}\|_2^2]$, i.e., $\EE[\|\bZ_{1}\|_2^2]\leq\Gamma$, where $\Gamma=2(1+1/m)(b+2G^2+d/\beta)$. Now assume that the conclusion holds for all iteration from $1$ to $k$, then for the $(k+1)$-th iteration, by \eqref{eq:2uuuu1} we have,
\begin{align}\label{eq:kkkk1}
    \EE[\|\bZ_{k+1}\|_2^2] 
    &=\EE[\|\bZ_{k}-\eta\tilde{\nabla}_{k}\|_2^2]+\frac{2\eta d}{\beta},
\end{align}
For the first term on the R.H.S of \eqref{eq:kkkk1} we have
\begin{align}\label{eq:2uuuu2}
\EE[\|\bZ_k-\eta\tilde{\nabla}_{k}\|_2^2]&=\EE[\|\bZ_k-\eta \nabla F_n(\bZ_k)\|_2^2] + 2\eta \EE \la \bZ_k-\eta \nabla F_n(\bZ_k), \nabla F_n(\bZ_k) - \tilde{\nabla}_{k} \ra\notag\\
&\quad\quad+ \eta^2 \EE[\|\nabla F_n(\bZ_k) -\tilde{\nabla}_{k}\|_2^2]\notag\\
&= \underbrace{\EE[\|\bZ_k-\eta \nabla F_n(\bZ_k)\|_2^2]}_{T_1}  + \underbrace{\eta^2 \EE[\|\nabla F_n(\bZ_k) -\tilde{\nabla}_{k}\|_2^2]}_{T_2},
\end{align}
where the second equality holds due to the fact that $\EE [\tilde{\nabla}_{k}]  = \nabla F_n(\bZ_k)$.
For term $T_1$, we can further bound it by
\begin{align*}
\EE[\|\bZ_k-\eta \nabla F_n(\bZ_k)\|_2^2]&=\EE[\|\bZ_k\|_2^2]-2\eta\EE[\langle\bZ_k,\nabla F_n(\bZ_k)\rangle]+\eta^2\EE[\|\nabla F_n(\bZ_k)\|_2^2]\\
&\leq \EE[\|\bZ_k\|_2^2]+2\eta(b-m\EE[\|\bZ_k\|_2^2])+2\eta^2(M^2\EE[\|\bZ_k\|_2^2]   +G^2)\\
&=(1-2\eta m+2\eta^2M^2)\EE[\|\bZ_k\|_2^2]  +2\eta b+2\eta^2G^2,
\end{align*}
where the inequality follows from Lemma \ref{lemma:gradient_bound} and triangle inequality.
For term $T_2$, by Lemma \ref{lemma:variance_svrg} we have
\begin{align*}
     \EE \| \nabla F_n(\bZ_k) - \tilde\nabla_{k} \|_2^2  \leq \frac{M^2(n-B)}{B(n-1)} \EE  \big\| \bZ_k - \tilde{\bZ}^{(s)}\big\|_2^2 \leq \frac{2M^2(n-B)}{B(n-1)} \Big(\EE  \big\| \bZ_k  \big\|_2^2 + \EE  \big\|  \tilde{\bZ}^{(s)}\big\|_2^2\Big).
\end{align*}
Submit the above bound back into \eqref{eq:2uuuu1} we have 
\begin{align}\label{eq:kkkk2}
\EE[\|\bZ_{k+1}\|_2^2]&\leq \bigg(1-2\eta m+2\eta^2M^2\Big(1 + \frac{n-B}{B(n-1)}\Big) \bigg)\EE[\|\bZ_k\|_2^2]\notag\\
&\quad\quad+ \frac{2\eta^2M^2(n-B)}{B(n-1)} \EE  \big\|  \tilde{\bZ}^{(s)}\big\|_2^2 + 2\eta b+2\eta^2G^2+\frac{2\eta d}{\beta}.
\end{align}
Note that by assumption we have $\EE  \big\|  {\bZ_j}\big\|_2^2 \leq \Gamma$ for all $j = 1,\dots,k$ where $\Gamma = 2\big(1+1/m\big)\big(b+2G^2+d/\beta\big)$, thus \eqref{eq:kkkk2} can be further bounded as:
\begin{align}\label{eq:kkkk3}
\EE[\|\bZ_{k+1}\|_2^2]\leq \underbrace{\bigg(1-2\eta m+2\eta^2M^2\Big(1 + \frac{2(n-B)}{B(n-1)}\Big) \bigg)}_{C_\lambda}\Gamma + 2\eta b+2\eta^2G^2+\frac{2\eta d}{\beta}.
\end{align}
For sufficient small $\eta$ that satisfies
$$\eta \leq \min\Bigg(1,\frac{m}{2M^2\big(1 + 2(n-B)/(B(n-1))\big)}\Bigg),$$ 
there are only two cases we need to take into account:\\
If $C_\lambda\leq 0$, then from \eqref{eq:kkkk3} we have
\begin{align}\label{eq:kkkk4}
    \EE[\|\bZ_{k+1}\|_2^2]&\leq 2\eta b+2\eta^2G^2+\frac{2\eta d}{\beta}\leq 2\bigg(b+G^2+\frac{d}{\beta}\bigg).
\end{align}
If $0<C_\lambda\leq 1$, then iterate \eqref{eq:kkkk3} and we have
\begin{align}\label{eq:kkkk5}
    \EE[\|\bZ_{k+1}\|_2^2]&\leq C_\lambda^{k+1}\|\bZ_0\|_2^2+\frac{\eta b+\eta^2G^2+\frac{\eta d}{\beta}}{\eta m-\eta^2M^2\Big(1 + \frac{2(n-B)}{B(n-1)}\Big) }\leq \frac{2}{m}\bigg(b+G^2+\frac{d}{\beta}\bigg).
\end{align}
Combining \eqref{eq:kkkk4} and \eqref{eq:kkkk5}, we have
\begin{align*}
    \EE[\|\bZ_{k+1}\|_2^2]&\leq 2\bigg(1+\frac{1}{m}\bigg)\bigg(b+2G^2+\frac{d}{\beta}\bigg).
\end{align*}
Thus we show that when $\EE[\|\bZ_{j}\|_2^2], j= 1,\dots,k$ are bounded, $\EE[\|\bZ_{k+1}\|_2^2]$ is also bounded. By mathematical induction we complete the proof.
\end{proof}

\subsection{Proof of Lemma \ref{lemma:L2_bound_exp_Xk}}
\begin{proof}
We have the following equation according to the update of GLD in \eqref{eq:gld},
\begin{align}\label{eq:proofc9_1}
\EE[\exp\big(\|\bX_{k+1}\|_2^2)]&=\EE\exp\bigg(\Big\|\bX_k-\eta\nabla F_n(\bX_k)+\sqrt{\frac{2\eta}{\beta}}\bepsilon_k\Big\|_2^2\bigg)\nonumber\\
&=\EE\exp\bigg(\|\bX_k-\eta\nabla F_n(\bX_k)\|_2^2+\sqrt{\frac{8\eta}{\beta}}\la\bX_k-\eta\nabla F_n(\bX_k),\bepsilon_k\ra+\frac{2\eta}{\beta}\|\bepsilon_k\|_2^2\bigg).
\end{align}
Let $H(\xb)=\exp(\|\xb\|_2^2)$, we have $\EE [H(\bX_{k+1})]=\EE_{\bX_k}[\EE[H(\bX_{k+1})|\bX_k]]$. Thus we can first compute the conditional expectation on the R.H.S of \eqref{eq:proofc9_1} given $\bX_k$, then compute the expectation with respect to $\bX_k$. Note that $\bepsilon_k$ follows standard multivariate normal distribution, i.e., $\bepsilon_k\sim N(\boldsymbol{0},\Ib_{d\times d})$. Then it can be shown that
\begin{align*}
&\EE\bigg[\exp\bigg(\sqrt{\frac{8\eta}{\beta}}\la\bX_k-\eta\nabla F_n(\bX_k),\bepsilon_k\ra+\frac{2\eta}{\beta}\|\bepsilon_k\|_2^2\bigg)\bigg|\bX_k\bigg]\\
&=\frac{1}{\big(1-4\eta/\beta\big)^{d/2}}\exp\bigg(\frac{4\eta}{\beta-4\eta}\|\bX_k-\eta\nabla F_n(\bX_k)\|_2^2\bigg)
\end{align*}
holds as long as $\beta>4\eta$. Plugging the above equation into \eqref{eq:proofc9_1}, we have
\begin{align}\label{eq:proofc9_2}
\EE[H(\bX_{k+1})]&=\frac{1}{\big(1-4\eta/\beta\big)^{d/2}}\EE_{\bX_k}\bigg[\exp\bigg(\frac{\beta}{\beta-4\eta}\|\bX_k-\eta\nabla F_n(\bX_k)\|_2^2\bigg)\bigg].
\end{align}
Note that by Assumption \ref{assump:dissipative} and Lemma \ref{lemma:gradient_bound} we have
\begin{align*}
&\EE_{\bX_k}\exp\bigg(\frac{\beta}{\beta-4\eta}\|\bX_k-\eta\nabla F_n(\bX)\|_2^2\bigg)\\
&=\EE_{\bX_k}\exp\bigg(\frac{\beta}{\beta-4\eta}\big(\|\bX_k\|_2^2-2\eta\la\bX_k,\nabla F_n(\bX_k)\ra+\eta^2\|\nabla F_n(\bX_k)\|_2^2\big)\bigg)\\
&\le\EE_{\bX_k}\exp\bigg(\frac{\beta}{\beta-4\eta}\big(\|\bX_k\|_2^2-2\eta(m\|\bX_k\|_2^2-b)+2\eta^2(M^2\|\bX_k\|_2^2+G^2)\big)\bigg)\\
&=\EE_{\bX_k}\exp\bigg(\frac{\beta}{\beta-4\eta}\big((1-2\eta m+2\eta^2M^2)\|\bX_k\|_2^2+2b\eta+2\eta^2G^2\big)\bigg).
\end{align*}
Consider sufficiently small $\eta$ such that $\eta<m/M^2$. Then for $\beta$ satisfying $\beta\geq\max\{2/(m-M^2\eta),4\eta\}$, we have $\beta(1-2\eta m+2\eta^2M^2)/(\beta-4\eta)\leq 1$. Therefore, the above expectation can be upper bounded by
\begin{align*}
\EE_{\bX_k}\exp\bigg(\frac{\beta}{\beta-4\eta}\|\bX_k-\eta\nabla F_n(\bX)\|_2^2\bigg)
&\le\exp\bigg(\frac{2\eta\beta(b+\eta G^2)}{\beta-4\eta}\bigg)\EE[H(\bX_k)].
\end{align*}
Substituting the above inequality into \eqref{eq:proofc9_2}, it follows that
\begin{align*}
\EE [H(\bX_{k+1})]&\le\frac{1}{(1-4\eta/\beta)^{d/2}}\exp\bigg(\frac{2\eta\beta(b+\eta G^2)}{\beta-4\eta}\bigg)\EE[H(\bX_k)]\\
&\le \exp\bigg(\frac{2\eta(\beta b+\eta\beta G^2+d)}{\beta-4\eta}\bigg)\EE [H(\bX_k)],
\end{align*}
where we used the fact that $\log(1/(1-x))\leq x/(1-x)$ for $0<x<1$ and that
\begin{align*}
\log\bigg(\frac{1}{(1-4\eta/\beta)^{d/2}}\bigg)=\frac{d}{2}\log\bigg(\frac{1}{1-4\eta/\beta}\bigg)\le\frac{2d\eta/\beta}{1-4\eta/\beta}=\frac{2\eta d}{\beta-4\eta}.
\end{align*}
Then we are able to show by induction that
\begin{align*}
\EE[ H(\bX_{k})]\le \exp\bigg(\frac{2k\eta(\beta b+\eta\beta G^2+d)}{\beta-4\eta}\bigg)\EE[ H(\|\bX_0\|_2)],
\end{align*}
which immediately implies that
\begin{align*}
\log \EE[\exp(\|\bX_k\|_2^2)]\le\|\bX_0\|_2^2+\frac{2\beta(b+G^2)+2d}{\beta-4\eta}k\eta,
\end{align*}
where we assume that $\eta\le1$ and $\beta>4\eta$.
\end{proof}

\subsection{Proof of Lemma \ref{lemma:variance_svrg}}

\begin{proof}
Since by Algorithm \ref{alg:vr_sgld} we have $\tilde{\nabla}_{k}=(1/B)\sum_{i_k \in I_k} \big( \nabla f_{i_k}(\bZ_k)-\nabla f_{i_k}(\tilde{\bZ}^{(s)})+\nabla F_n(\tilde{\bZ}^{(s)}) \big)$, therefore,
\begin{align*} 
     \EE [\|\tilde\nabla_{k}   -\nabla F_n(\bZ_k)  \|_2^2 ]  &=  \EE  \bigg\| \frac{1}{B} \sum_{i_k \in I_k} \big(\nabla f_{i_k}(\bZ_k)-\nabla f_{i_k}(\tilde{\bZ}^{(s)})+\nabla F_n(\tilde{\bZ}^{(s)})- \nabla F_n(\bZ_k) \big)  \bigg\|_2^2.
\end{align*}
Let $\ub_i = \nabla F_n(\bZ_k) - \nabla F_n(\tilde{\bZ}^{(s)}) - \big(\nabla f_{i_k}(\bZ_k) - \nabla f_{i_k}(\tilde{\bZ}^{(s)}) \big)$. 
\begin{align}
    \EE \bigg\| \frac{1}{B}\sum_{i \in I_k} \ub_i(\xb) \bigg\|_2^2 &= \frac{1}{B^2} \EE \sum_{i \neq i' \in I_k} \ub_i(\xb)^\top \ub_{i'}(\xb) + \frac{1}{B}\EE\|\ub_i(\xb)\|_2^2\notag\\
    &= \frac{B-1}{Bn(n-1)} \sum_{i \neq i'} \ub_i(\xb)^\top \ub_{i'}(\xb) + \frac{1}{B}\EE\|\ub_i(\xb)\|_2^2\notag\\
    &= \frac{B-1}{Bn(n-1)} \sum_{i, i'} \ub_i(\xb)^\top \ub_{i'}(\xb) -\frac{B-1}{B(n-1)}\EE\|\ub_i(\xb)\|_2^2 + \frac{1}{B}\EE\|\ub_i(\xb)\|_2^2\notag\\
    &= \frac{n-B}{B(n-1)}\EE\|\ub_i(\xb)\|_2^2,
\end{align}
where the last equality is due to the fact that $1/n\sum_{i=1}^n \ub_i(\xb)  = 0$. 
Therefore, we have
\begin{align}\label{eq:j4444}
     \EE [\|\tilde\nabla_{k}   -\nabla F_n(\bZ_k)  \|_2^2 ]  &\leq  \frac{n-B}{B(n-1)} \EE \| \ub_i \|_2^2  \notag\\
     &=\frac{n-B}{B(n-1)}  \EE   \|\nabla f_{i_k}(\bZ_k)-\nabla f_{i_k}(\tilde{\bZ}) - \EE[\nabla f_{i_k}(\bZ_k)-\nabla f_{i_k}(\tilde{\bZ})]\|_2^2  \notag\\
     &\leq  \frac{n-B}{B(n-1)}  \EE   \|\nabla f_{i_k}(\bZ_k)-\nabla f_{i_k}(\tilde{\bZ})\|_2^2 \notag\\
     &\leq\frac{M^2(n-B)}{B(n-1)} \EE  \| \bZ_k - \tilde{\bZ}\|_2^2,
\end{align}
where the second inequality holds due to the fact that $\EE[\|\xb - \EE[\xb]\|_2^2] \leq \EE[\|\xb\|_2^2]$ and the last inequality follows from Assumption \ref{assump:smooth}. This completes the proof.
\end{proof}



\section{Proof of Auxiliary Lemmas in Appendix \ref{sec:proof_of_aux_lemma}}
\subsection{Proof of Lemma \ref{lemma:gradient_bound}}
\begin{proof}
By Assumption \ref{assump:dissipative} we obtain
\begin{align*}
\la \xb^*, \nabla F_n(\xb^*) \ra \geq m\|\xb^*\|_2^2 -b.
\end{align*}
Note that $\xb^*$ is the minimizer for $F_n$, which implies that  $\nabla F_n(\xb^*)= \zero$ and threfore $\|\xb^*\|_2 \leq b/m$. By Assumption \ref{assump:smooth} we further have
\begin{align*}
 \|\nabla f_i(\xb)\|_2 \leq \|\nabla f_i(\xb^*)\|_2 + M\|\xb -\xb^*\|_2 \leq \|\nabla f_i(\xb^*)\|_2 + \frac{bM}{m} + M\|\xb\|_2.
\end{align*}
The proof is completed by setting $G = \max_{i=1,\ldots,n}\{\|\nabla 
f_i(\xb^*)\|_2\} + bM/m$.
\end{proof}

\end{document}